\documentclass{article}

\PassOptionsToPackage{numbers}{natbib}
 \usepackage[preprint]{neurips_2026}
\usepackage[utf8]{inputenc} 
\usepackage[T1]{fontenc}    
\usepackage{hyperref}       
\usepackage{url}            
\usepackage{booktabs}       
\usepackage{amsfonts}       
\usepackage{nicefrac}       
\usepackage{microtype}      
\usepackage{xcolor}         
\usepackage{amsmath}
\usepackage{wrapfig}      
\usepackage{booktabs}     
\usepackage[table]{xcolor}
\usepackage{algorithm}
\usepackage{algorithmic}
\usepackage{graphicx}
\usepackage{multirow}
\usepackage{titletoc}

\usepackage{bbding}
\usepackage{pifont}
\usepackage{wasysym}

\title{Physical Fidelity Reconstruction via Improved Consistency-Distilled Flow Matching for Dynamical Systems}

\author{%
  Sicheng Ma\thanks{Equal Contribution.}\\
  Department of Physics\\
  University of Cambridge\\
  \And
  Tianyue Yang\protect\footnotemark[1] \\
  The Center for Computational Science\\
  University College London\\
  \And
  Xiuzhe Wu \\
  Department of Computer Science\\
  University College London\\
  \And
  Xiao Xue\thanks{Corresponding Author.}\\
  The Center for Computational Science\\
  University College London\\
  \texttt{x.xue@ucl.ac.uk}\\
}

\begin{document}

\maketitle

\begin{abstract}
Reconstructing high-fidelity flow fields from low-fidelity observations is a central problem in scientific machine learning, yet recent diffusion and flow-matching models typically rely on iterative sampling, making them costly for latency-sensitive workflows such as ensemble forecasting, real-time visualization, and simulation-in-the-loop inference. We study whether a high-fidelity flow-matching generative model can be compressed into a compact one-step model for fast scientific flow reconstruction. Our approach distills an optimal-transport flow-matching teacher into a one-step consistency model. Low-fidelity observations are incorporated at inference by initializing the generative trajectory from a noised observation along the transport path, allowing an unconditional high-fidelity flow model to perform conditional reconstruction without retraining the teacher. We evaluate this distillation strategy on three fluid benchmarks, Smoke Buoyancy, Turbulent Channel Flow, and Kolmogorov Flow, using coarse-to-fine reconstruction as a controlled testbed at field sizes up to $256 \times 256$. Across these settings, the distilled student retains similar performance of the teacher's model on spectrum metrics, while using roughly half as many parameters and achieving a $12\times$ inference speedup over the flow-matching teacher. Under the same training budget, the distilled student also outperforms a one-step consistency model trained directly from scratch by $23.1\%$ in SSIM, showing that teacher distillation improves training efficiency rather than merely accelerating sampling. These results suggest a promising route for turning future high-capacity scientific generative models into compact reconstruction models that are faster to train, cheaper to run, and easier to deploy.
\end{abstract}

\section{Introduction}

The high-fidelity simulation of physical systems such as turbulent flows, atmospheric dynamics, and combustion processes is central to modern scientific computing. These systems are governed by nonlinear partial differential equations whose solutions exhibit rich multi-scale structure, with small-scale features playing a critical role in energy transfer, dissipation, and long-term system evolution~\cite{pope2000turbulent}. Direct Numerical Simulation (DNS)~\cite{lee2015direct} resolves these features faithfully but at computational costs that are prohibitive for many practical workflows such as ensemble forecasting, parameter sweeps, real-time monitoring, and uncertainty quantification routinely require thousands or millions of forward evaluations, far beyond what DNS can support. Approximation methods such as Reynolds-Averaged Navier-Stokes (RANS)~\cite{alfonsi2009reynolds} and Large Eddy Simulation (LES)~\cite{piomelli1999large} reduce cost by filtering or modelling fine-scale dynamics, but at the expense of the very small-scale fidelity that often matters most.

A practical alternative is \emph{physical fidelity enhancement}, also referred to as super-resolution or statistical downscaling: given a coarse, cheaply-computed low-resolution field, recover a high-resolution counterpart consistent with the underlying physics. The mapping is fundamentally ill-posed as multiple physically plausible high-resolution states are consistent with any low-resolution observation, which makes generative modelling a natural fit~\cite{wan2023debias}. Diffusion-based approaches~\cite{ho2020denoising, song2020score, song2019generative} and, more recently, Flow Matching (FM)~\cite{lipman2022flow, albergo2023stochastic} have produced strong results on this class of problem~\cite{lin2024diffbir, wang2024sinsr}, yielding samples that respect the energy spectra of the target dynamics and capture multi-modal posteriors over fine-scale detail~\cite{xue2026uni, yang2026meno, shu2023physics, oommen2025integrating, lin2026decoupled}.

Despite their fidelity, these models are intrinsically multi-step at inference: producing a single high-resolution field requires several Neural Function Evaluations (NFEs) along an iterative denoising or integration trajectory. The cost compounds when many samples are needed, which is precisely the regime in which generative super-resolution would be most useful: large-ensemble forecasting, simulation-in-the-loop pipelines, real-time visualisation. This inference cost has become the binding constraint on deploying generative super-resolution in scientific workflows, and it cannot be circumvented by simply scaling hardware because the latency of a multi-step trajectory is irreducible at the algorithmic level.

Consistency models~\cite{song2023consistency,song2023improved} and their simplified variant (sCM)~\cite{lu2024simplifying} address this bottleneck by learning a self-consistent mapping that collapses an entire generative trajectory into a single forward pass. Recent work by \citet{chen2025sanasprint} demonstrates that a pretrained FM teacher can be distilled into a one-step sCM student via TrigFlow parameterisation and Jacobian-vector-product based tangent estimation, achieving teacher-level quality at a fraction of the inference cost. This framework has been validated extensively in the natural-image domain. Whether it transfers to scientific fields with their power-law spectra, conservation structure, and multi-scale coupling has not been demonstrated.

In this work we present the first such demonstration. We distil an FM teacher into a TrigFlow consistency student for physical fidelity enhancement of two-dimensional fluid systems, with both teacher and student trained fully unconditionally on the high-resolution distribution. The low-resolution observation is incorporated only at inference, by initializing the trajectory from a noised version of the coarse field at an intermediate point along the OT path using the FM path insertion mechanism~\cite{meng2022sdedit, xu2025fast, kawar2022denoising} that carries through cleanly to the one-step consistency student.

\paragraph{Contributions.}
\begin{itemize}
  \item We present the first demonstration of one-step consistency distillation from a flow-matching teacher for physical fidelity enhancement of fluid dynamical systems, adapting the sCM/TrigFlow framework~\cite{lu2024simplifying,chen2025sanasprint} to scientific super-resolution.
  \item Across three fluid benchmarks, our 15M-parameter student matches a substantial fraction of the 30M-parameter teacher's improvement over low-resolution baselines on both pointwise and spectral metrics, while reducing inference to a single network evaluation and achieving a $12\times$ wall-clock speedup over the teacher.
  \item We provide a systematic benchmark across resolutions ($128\times128$ to $256\times256$) and flow regimes (Smoke Buoyancy, Turbulent Channel Flow, 2D Kolmogorov Flow), establishing reference results for one-step generative super-resolution in fluid dynamics.
\end{itemize}

\section{Related Works}
\paragraph{Diffusion models for physical fidelity reconstruction.}
Diffusion-based generative models have long been adapted for physically faithful reconstruction~\cite{schiodt2026generative, shu2023physics, xue2026uni}. They have also demonstrated superior performance over deterministic reconstruction algorithms~\cite{shu2023physics}, largely because their distribution-matching nature enables better preservation of physical metrics~\cite{shu2023physics, xue2026uni, yang2026meno}. \citet{shu2023physics} extensively studied the application of denoising diffusion probabilistic models (DDPMs)~\cite{ho2020denoising} and denoising diffusion implicit models (DDIMs)~\cite{song2022denoisingdiffusionimplicitmodels}, showing that diffusion models consistently outperform deterministic baselines on super-resolution and sparse reconstruction tasks. Flow matching, as an alternative formulation of diffusion models~\cite{lipman2022flow, albergo2023stochastic}, has also been widely adopted for generative vision reconstruction and editing tasks~\cite{kulikov2025flowedit, delefosse2026super, hadzic2026restora, fallah2025rareflow}, including applications in Earth systems such as weather forecasting~\cite{fallah2025rareflow, delefosse2026super} and fluid dynamics~\cite{schiodt2026generative}. However, these models are intrinsically multi-step and incur substantial inference costs as the number of generated frames increases, limiting their applicability in real-world scenarios~\cite{yang2026meno}. Although one-step generative models, such as consistency models~\cite{song2023consistency} and shortcut diffusion~\cite{frans2024one}, have been introduced to address this limitation, their application to physical systems remains largely unexplored.

\paragraph{Diffusion distillation.} To mitigate the inference cost of standard DDPM and flow matching models, distillation methods have been proposed to reduce the number of inference steps required by multi-step generative models. Diffusion distillation methods such as Distribution Matching Distillation (DMD)~\cite{yin2024one, yin2024improved} and Consistency Trajectory Models (CTM)~\cite{kim2023consistency} involve the joint training of student and adversarial networks, which can impose substantial training overhead. A simple yet effective alternative is consistency distillation~\cite{song2023consistency, lu2024simplifying, yang2024consistency}, which distils FM-style models by enforcing consistency along generation trajectories. In this study, we focus on this formulation to achieve efficient and effective distillation.

\section{Methodology}
\label{sec:methodology}

\begin{figure}[t]
  \centering
  \includegraphics[width=\linewidth]{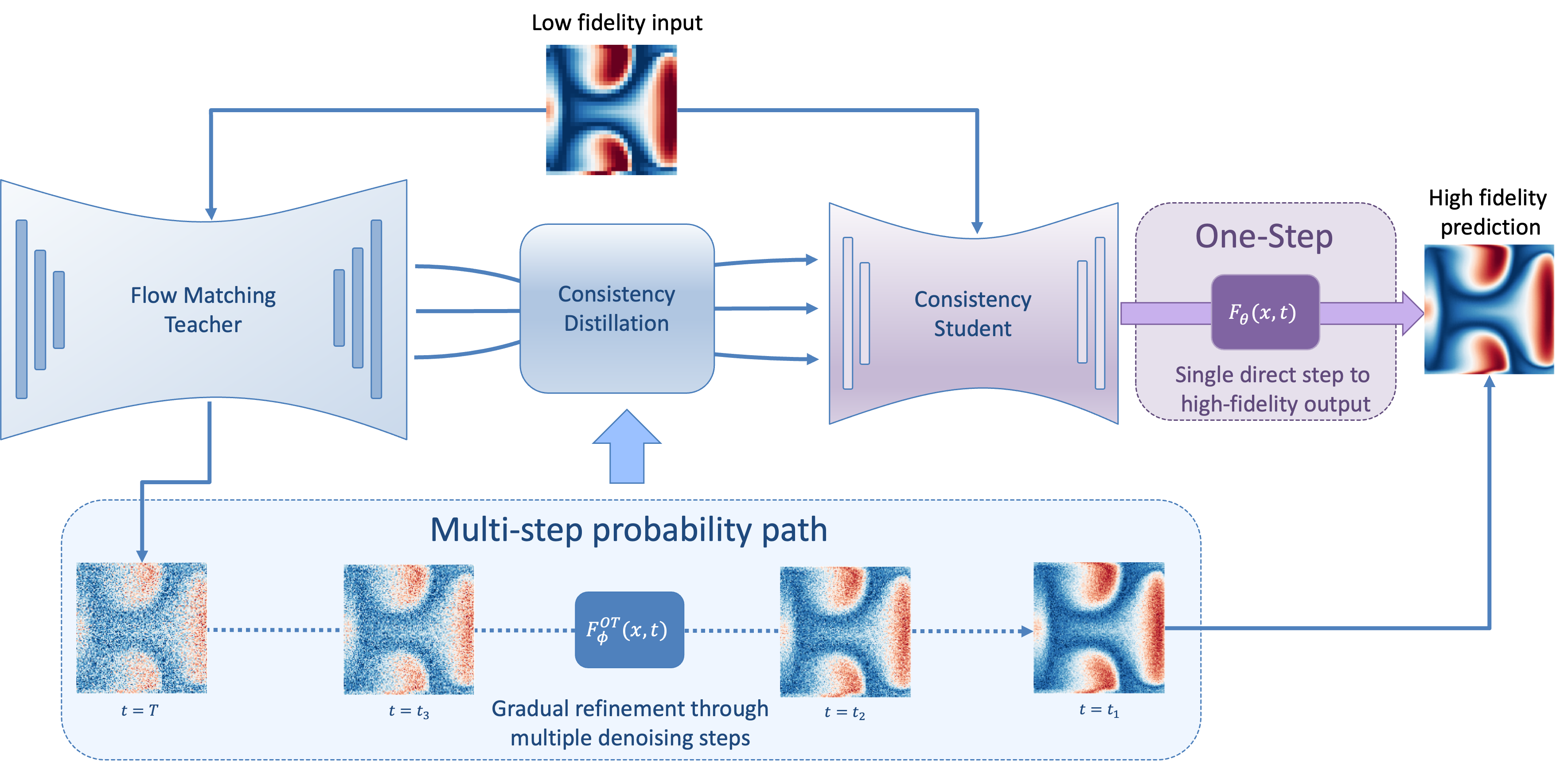}
  \caption{\textbf{Overview.} \textit{Left:} Our pipeline distills a multi-step OT-FM teacher (5-step RK5, 30 NFEs) into a one-step TrigFlow consistency student via sCM distillation. Both models are trained unconditionally; the low-resolution observation is incorporated at inference by initializing the trajectory at an intermediate time $\tau$ along the OT path.}
  \label{fig:overview}
\end{figure}

Figure~\ref{fig:overview} summarizes our system. We train an unconditional OT-FM teacher on the high-resolution distribution (Section~\ref{sec:fm}), distill it into a one-step TrigFlow consistency student via sCD (Section~\ref{sec:distillation}), and incorporate the low-resolution observation purely at inference by initializing the trajectory along the OT path, reducing the full pipeline to a single forward pass through the student (Section~\ref{sec:conditioning}).

\subsection{Problem Setup}
\label{sec:problem}

We address physical fidelity enhancement of two-dimensional fluid fields. Let $\mathbf{x}_{\text{HR}} \in \mathbb{R}^{H \times W \times C}$ denote a high-resolution snapshot of a fluid field with $C$ channels (e.g.\ velocity components, density), and let $\mathbf{x}_{\text{LR}} \in \mathbb{R}^{h \times w \times C}$ denote a corresponding low-resolution observation obtained by stride-$s$ subsampling of $\mathbf{x}_{\text{HR}}$, with $H = sh$ and $W = sw$. The goal is to recover a high-resolution sample $\hat{\mathbf{x}}_{\text{HR}}$ consistent with the underlying physics, conditioned on $\mathbf{x}_{\text{LR}}$.

The mapping is fundamentally one-to-many: many physically plausible high-resolution fields are consistent with any given low-resolution observation, since the subsampling discards information at scales smaller than $s$. We therefore model $p(\mathbf{x}_{\text{HR}} \mid \mathbf{x}_{\text{LR}})$ as a conditional distribution and draw samples from it via a generative trajectory.

Throughout, we represent the low-resolution input on the high-resolution grid via nearest-neighbour upsampling, denoted $\mathbf{x}_{\text{LR}}^{\uparrow} \in \mathbb{R}^{H \times W \times C}$, so that all flow trajectories operate in the same ambient space.

\subsection{Flow Matching}
\label{sec:fm}
Flow Matching~\cite{lipman2022flow, albergo2023stochastic} learns a continuous-time velocity field $\mathbf{v}_\phi(\mathbf{z}, t)$ that transports a Gaussian base distribution into the data distribution $p_{\text{data}}$ along a time-indexed density path $\{p_t\}_{t \in [0,1]}$. The path is specified through a conditional coupling $\mathbf{z}_t = a_t\,\mathbf{x} + b_t\,\boldsymbol{\epsilon}$ with $\mathbf{x} \sim p_{\text{data}}$, $\boldsymbol{\epsilon} \sim \mathcal{N}(\mathbf{0}, \mathbf{I})$, and $\mathbf{v}_\phi$ is trained to regress the corresponding conditional velocity $\mathrm{d}\mathbf{z}_t/\mathrm{d}t$ via the conditional flow-matching loss~\cite{lipman2022flow}. We adopt the linear (optimal-transport) schedule $a_t = 1 - t$, $b_t = t$, which yields straight-line trajectories
\begin{equation}
\mathbf{z}_t = (1 - t)\,\mathbf{x} + t\,\boldsymbol{\epsilon}, \qquad t \in [0, 1],
\label{eq:ot_path}
\end{equation}
and a constant conditional velocity $\boldsymbol{\epsilon} - \mathbf{x}$~\cite{lipman2022flow}. Our teacher is trained under this OT parameterization. The full FM objective, its conditional reformulation (CFM), and the OT-path derivation are reproduced in Appendix~\ref{app:fm}.

\subsection{Consistency Distillation}
\label{sec:distillation}

\paragraph{Consistency models.}
A consistency model~\cite{song2023consistency,song2023improved} replaces iterative integration with a single network $f_\theta(\mathbf{z}_t, t)$ trained to map any point along a generative trajectory directly to that trajectory's endpoint, enforcing the self-consistency property
\begin{equation}
f_\theta(\mathbf{z}_t, t) = f_\theta(\mathbf{z}_{t'}, t') \qquad \text{for all } (t, t') \text{ along the same trajectory.}
\label{eq:self_consistency}
\end{equation}
Once trained, generation reduces to a single forward pass, regardless of where on the trajectory the input lies.

\paragraph{Continuous-time consistency distillation.}
We adopt the simplified continuous-time consistency distillation (sCD) framework of \citet{lu2024simplifying}, which differentiates the self-consistency identity directly rather than approximating it with finite differences over discrete trajectory pairs. Concretely, the student's velocity is regressed onto a target constructed from an EMA copy of the student and the tangent of the consistency function along the trajectory. The tangent term is computed exactly using a Jacobian--vector product (JVP), requiring only a single additional forward-mode pass, with the trajectory direction provided by the teacher. We provide the full sCD loss and JVP-based tangent estimator in Appendix~\ref{app:cm}, and the per-step distillation algorithm in Appendix~\ref{app:algorithms}. The sCD framework is derived under a trigonometric coupling, referred to as TrigFlow. Following the lossless OT--TrigFlow equivalence of \citet{chen2025sanasprint}, we query our pretrained OT-FM teacher at TrigFlow timesteps, enabling it to provide consistent supervision to the student without retraining. The required transformations are given in Appendix~\ref{app:trigflow}.

\subsection{Conditioning and Inference}
\label{sec:conditioning}

Both the FM teacher and the sCM student are trained \emph{unconditionally} on the high-resolution distribution $p(\mathbf{x}_{\text{HR}})$. The low-resolution observation is incorporated only at inference by exploiting the structure of the OT path: we treat the upsampled LR field $\mathbf{x}_{\text{LR}}^{\uparrow}$ as an approximation to a high-resolution sample that lies on the OT trajectory at some intermediate time $\tau \in (0, 1)$, and initialize the inference trajectory at that point,
\begin{equation}
\mathbf{z}_\tau = (1 - \tau)\,\mathbf{x}_{\text{LR}}^{\uparrow} + \tau\,\boldsymbol{\epsilon}, \qquad \boldsymbol{\epsilon} \sim \mathcal{N}(\mathbf{0}, \mathbf{I}).
\label{eq:noise_injection}
\end{equation}
This is precisely the affine interpolation of Eq.~\ref{eq:ot_path} that the velocity field was trained on, so the initialization lies on-manifold for $\mathbf{v}_\phi$ at time $\tau$; the construction is the OT-path flow-matching analog of SDEdit~\cite{meng2022sdedit}. The hyperparameter $\tau$ trades off LR fidelity against high-resolution realism, small $\tau$ stays close to the LR input but adds little fine-scale structure, large $\tau$ erases the LR information, and within a wide plateau the choice is largely cosmetic (Appendix~\ref{app:tau_sweep}).

At test time the FM teacher integrates the unconditional velocity field from $t = \tau$ to $t = 0$ with a fifth-order Runge--Kutta solver (5 NFEs). The sCM student needs no integration: it maps any on-trajectory point directly to the endpoint in a single forward pass, so $\hat{\mathbf{x}}_{\text{HR}} = f_\theta(\mathbf{z}_\tau, \tau_{\text{Trig}})$ with $\tau$ expressed in TrigFlow time. The student is operated at a shallower noise injection ($\tau \approx 0.3$) than the multi-step teacher ($\tau \approx 0.6$), reflecting its reduced capacity to refine deeply noised inputs in a single step. The full inference pipeline thus reduces to one nearest-neighbour upsample, one noise sample, and one network evaluation.

\section{Experiments}
\label{sec:experiments}
We evaluate the proposed method on three two-dimensional fluid benchmarks spanning different resolutions and flow regimes.
Our central question is whether a one-step consistency student distilled from an OT-FM teacher can recover comparable physical fidelity at substantially lower inference cost.
All experiments follow the unconditional training setup in Section~\ref{sec:methodology}; task conditioning is introduced only at inference through trajectory initialization at a dataset-specific noising time $\tau$.

\paragraph{Benchmarks}
We consider three datasets:
\textbf{Smoke Buoyancy} ($32{\to}128$; buoyancy-driven transport; 173{,}824 snapshots),
\textbf{Turbulent Channel Flow (TCF)} ($64{\to}192$; anisotropic near-wall turbulence; 38{,}080 snapshots),
and \textbf{Kolmogorov Flow} ($64{\to}256$; two-dimensional periodic incompressible turbulence; 12{,}800 snapshots).
Appendix~\ref{app:datasets} provides full dataset statistics, simulation provenance, and preprocessing details.
Together, these benchmarks test whether the method generalizes across distinct fluid regimes and super-resolution scales rather than being tuned to a single setting.

\paragraph{Baselines}
We compare against four baselines.
The \textbf{LR baseline} is the nearest-neighbour upsampled low-resolution input and serves as a no-model reference.
The \textbf{FM teacher} is the multi-step OT-FM model used for distillation, evaluated by default with a 5-step RK5 ODE solver.
The \textbf{diffusion baseline} is a separately trained DDPM-style model, providing a standard multi-step generative reference.
Finally, \textbf{sCM trained from scratch} uses the same student architecture and consistency objective as our distilled model but receives no teacher supervision, isolating the effect of distillation.

\paragraph{Metrics}
We evaluate fidelity using both pointwise and spectral metrics.
Pointwise reconstruction is measured by relative $L_2$ error (R$L_2$) and Structural Similarity Index Measure (SSIM)~\cite{wang2004image} against the ground-truth high-resolution field.
Spectral fidelity is measured by Power Spectral Density Discrepancy (PSDD), the radially averaged deviation between predicted and ground-truth power spectra.
Because pointwise metrics can favor overly smooth reconstructions, PSDD is essential for assessing whether a model preserves the multi-scale energy structure of fluid fields.
Inference cost is reported as wall-clock time per frame on a single NVIDIA GPU, averaged over 100 forward passes after 10 warm-up passes.
All metrics are computed on the held-out 10\% validation split.

Across all benchmarks, the FM teacher's Euler, Heun, and RK5 solvers produce near-identical reconstruction metrics, indicating that solver order is not the limiting factor.
We therefore focus the per-dataset discussion on the 5-step RK5 teacher.

\subsection{Smoke Buoyancy}
\begin{table}[!h]
\centering
\caption{Smoke Buoyancy (32\,$\rightarrow$\,128) super-resolution metrics. The sCM-distilled model achieves comparable reconstruction quality to the flow matching teacher with roughly half the parameters and a single function evaluation. Diffusion rows compare DDIM and DPM-Solver++ (DPM++) at 30 reverse steps; flow-matching rows compare Euler, Heun, and the 5-step RK5 used by the teacher in the rest of the paper.}

\label{tab:smoke_metrics}
\setlength{\tabcolsep}{8pt}
\begin{tabular}{l|c|ccc}
\toprule
\textbf{Model} & \textbf{NFE} & R$L_2$ $\downarrow$ & SSIM $\uparrow$ & PSDD $\downarrow$ \\
\midrule
\rowcolor{gray!15} \multicolumn{5}{l}{\textit{Diffusion (multi-step)}}\\
DDIM          & 30 & 0.145 & 0.690 & $4.35\times10^{-2}$ \\
DPM++         & 30 & 0.156 & 0.615 & $4.38\times10^{-2}$ \\
\rowcolor{gray!15} \multicolumn{5}{l}{\textit{Flow Matching teacher (multi-step)}}\\
Euler         & 5 & 0.162 & 0.657 & $6.61\times10^{-2}$ \\
Heun          & 10 & 0.168 & 0.614 & $6.60\times10^{-2}$ \\
RK5           & 30 & 0.168 & 0.628 & $6.60\times10^{-2}$ \\
\rowcolor{gray!15} \multicolumn{5}{l}{\textit{Consistency Models (1-step)}}\\
sCM (trained)    &   1      & 0.19 & 0.54 & $1.75\times10^{-1}$ \\
\textbf{sCM (distilled)}& 1 & \textbf{0.158} & \textbf{0.665} & $\mathbf{4.23\times10^{-2}}$ \\
\midrule
\multicolumn{2}{c|}{$\Delta$ vs.\ RK5 (\%)} & $-5.9$ & $+5.9$ & $-35.9$ \\
\bottomrule
\end{tabular}
\end{table}

Table~\ref{tab:smoke_metrics} reports reconstruction metrics on the Smoke Buoyancy dataset ($32{\to}128$).
The 30-step diffusion baselines, DDIM and DPM++, use a substantially smaller backbone ($3.7$\,M parameters, approximately $9{\times}$ smaller than the FM teacher) yet achieve slightly stronger pointwise reconstruction scores than the FM teacher itself. This provides useful context for the wall-clock comparison in Table~\ref{tab:inference_consolidated}.
Despite using only a single function evaluation, the distilled sCM model (15.9\,M parameters, NFE$=1$) outperforms the RK5 FM teacher on all reported metrics, improving R$L_2$ by $5.9\%$, SSIM by $5.9\%$, and PSDD by $35.9\%$.
In contrast, the from-scratch sCM ablation, which uses the same student architecture and training budget but receives no teacher supervision, performs substantially worse on both pixel metrics and spectral metrics.
This gap isolates teacher distillation, rather than the consistency objective alone, as the key factor enabling accurate one-step generation on this task.

\begin{figure}[!h]
  \centering
  \includegraphics[width=\linewidth]{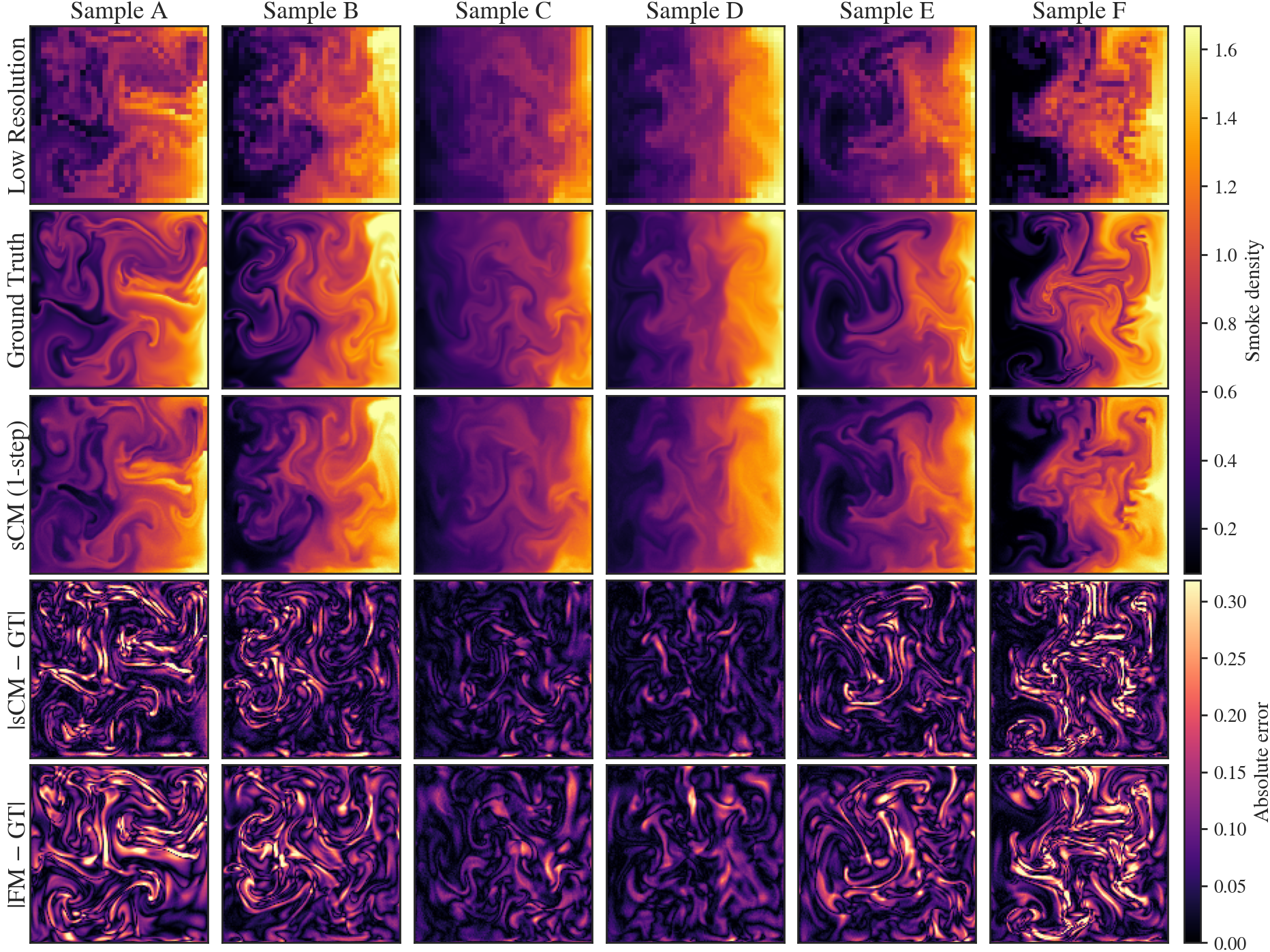}
  \caption{\textbf{Smoke Buoyancy super-resolution samples} ($32{\to}128$). Six independently super-resolved frames (Samples A--F) drawn from across the dataset; columns are i.i.d.\ reconstructions of distinct frames. \textit{Row 1:} nearest-neighbour-upsampled low-resolution input. \textit{Row 2:} high-resolution ground truth. \textit{Row 3:} one-step sCM student reconstructions, which preserve the buoyant plume's vortical roll-ups, edge sheets, and fine-scale fingering. \textit{Rows 4--5:} per-pixel absolute residual against ground truth for the sCM student and the five-step OT-FM teacher, plotted on a shared error colorbar. The student's residual is comparable to and, on several samples, lower than the teacher's despite using a single network evaluation versus thirty.}
  \label{fig:smoke_trajectory}
\end{figure}

Figure~\ref{fig:smoke_trajectory} confirms the metric numbers visually: across eight i.i.d.\ samples spanning a range of plume morphologies, the one-step student is nearly indistinguishable from the ground truth at this resolution, and its absolute-error map closely tracks the teacher's both in spatial localisation (concentrated along the plume's steep density gradients) and in magnitude. The headline message is that the distillation does not collapse fine-scale structure: vortical features at the diffusive front are recovered in a single forward pass.

\subsection{Turbulent Channel Flow}
\begin{table}[!h]
\centering
\caption{Turbulent Channel Flow (64\,$\rightarrow$\,192) super-resolution metrics. The sCM-distilled model preserves the reconstruction fidelity of the flow matching teacher while operating at NFE\,$=$\,1 with approximately half the parameter count. Diffusion rows compare DDIM and DPM-Solver++ at 30 reverse steps; flow-matching rows compare Euler, Heun, and the 5-step RK5 used by the teacher in the rest of the paper.}
\label{tab:tcf_metrics}
\setlength{\tabcolsep}{8pt}
\begin{tabular}{l|c|ccc}
\toprule
\textbf{Model} & \textbf{NFE} & R$L_2$ $\downarrow$ & SSIM $\uparrow$ & PSDD $\downarrow$ \\
\midrule
\rowcolor{gray!15} \multicolumn{5}{l}{\textit{Diffusion (multi-step)}}\\
DDIM          & 30 & 0.0355 & 0.9611 & $2.72\times10^{-3}$ \\
DPM++         & 30 & 0.0192 & 0.9832 & $7.36\times10^{-4}$ \\
\rowcolor{gray!15} \multicolumn{5}{l}{\textit{Flow Matching teacher (multi-step)}}\\
Euler         & 5 & 0.0056 & 0.9984 & $2.94\times10^{-4}$ \\
Heun          & 10 & 0.0102 & 0.9709 & $1.98\times10^{-4}$ \\
RK5           & 30 & 0.0071 & 0.9946 & $3.58\times10^{-4}$ \\
\rowcolor{gray!15} \multicolumn{5}{l}{\textit{Consistency Models (1-step)}}\\
sCM (trained)           & 1 & 0.0479 & 0.7781 & $4.06\times10^{-2}$ \\
\textbf{sCM (distilled)} & 1 & \textbf{0.0173} & \textbf{0.9788} & $\mathbf{3.51\times10^{-3}}$ \\
\midrule
\multicolumn{2}{c|}{$\Delta$ vs.\ RK5 (\%)} & $+145.7$ & $-1.6$ & $+880$ \\
\bottomrule
\end{tabular}
\end{table}

Table~\ref{tab:tcf_metrics} reports results on Turbulent Channel Flow.
Among FM teacher solvers, Euler gives the best pointwise scores (R$L_2=0.0056$, SSIM$=0.998$), marginally exceeding RK5; at $N{=}16$ frames, this difference is small and likely reflects the smooth bulk velocity profile that dominates the $192^2$ field.
DPM-Solver++ substantially improves over DDIM (R$L_2$ $0.019$ vs.\ $0.036$; PSDD $7.4{\times}10^{-4}$ vs.\ $2.7{\times}10^{-3}$), suggesting that the higher-order solver better recovers near-wall gradients.
The distilled sCM remains close to the FM teacher in SSIM ($1.6\%$ gap) but incurs larger errors in R$L_2$ ($\sim\!2.4{\times}$) and PSDD (about one order of magnitude), making TCF the only benchmark where one-step distillation carries a notable spectral cost.
This gap is likely amplified by the dataset's narrow dynamic range and by the teacher's exceptionally low PSDD ($3.58{\times}10^{-4}$), which sets a high spectral baseline; the student's PSDD ($3.51{\times}10^{-3}$) is comparable to its values on the other datasets.
Qualitative trajectories in Appendix~\ref{app:qualitative} (Figure~\ref{fig:tcf_trajectory}) show that structural content is largely preserved, while residuals reveal the teacher's lower pointwise error.

\subsection{Kolmogorov Flow}
\begin{table}[!h]
\centering
\caption{Kolmogorov Flow (64\,$\rightarrow$\,256) super-resolution metrics. The sCM-distilled model matches the flow matching teacher across all metrics while requiring a single network evaluation and roughly half the parameters. On this benchmark the distilled student in fact \emph{improves over the FM teacher on every metric}, with a $-59.3\%$ reduction in spectral error (PSDD). Diffusion rows compare DDIM and DPM-Solver++ at 30 reverse steps; flow-matching rows compare Euler, Heun, and the 5-step RK5 used by the teacher in the rest of the paper.}
\label{tab:kf_metrics}
\setlength{\tabcolsep}{8pt}
\begin{tabular}{l|c|ccc}
\toprule
\textbf{Model} & \textbf{NFE} & R$L_2$ $\downarrow$ & SSIM $\uparrow$ & PSDD $\downarrow$ \\
\midrule
\rowcolor{gray!15} \multicolumn{5}{l}{\textit{Diffusion (multi-step)}}\\
DDIM          & 30 & 0.291 & 0.688 & $4.75\times10^{-2}$ \\
DPM++         & 30 & 0.293 & 0.681 & $4.93\times10^{-2}$ \\
\rowcolor{gray!15} \multicolumn{5}{l}{\textit{Flow Matching teacher (multi-step)}}\\
Euler         & 5 & 0.304 & 0.689 & $3.26\times10^{-2}$ \\
Heun          & 10 & 0.309 & 0.678 & $3.29\times10^{-2}$ \\
RK5           & 30 & 0.308 & 0.682 & $3.29\times10^{-2}$ \\
\rowcolor{gray!15} \multicolumn{5}{l}{\textit{Consistency Models (1-step)}}\\
sCM (trained)  &        1   & 0.282 & 0.723 & $5.27\times10^{-2}$ \\
\textbf{sCM (distilled)} & 1 & \textbf{0.287} & \textbf{0.717} & $\mathbf{1.34\times10^{-2}}$ \\
\midrule
\multicolumn{2}{c|}{$\Delta$ vs.\ RK5 (\%)} & $-6.8$ & $+5.1$ & $-59.3$ \\
\bottomrule
\end{tabular}
\end{table}

Table~\ref{tab:kf_metrics} reports results on Kolmogorov Flow, the highest-resolution and most challenging benchmark in our suite ($256^2$).
The distilled sCM outperforms all multi-step baselines: relative to the RK5 FM teacher, it reduces R$L_2$ by $6.8\%$, increases SSIM by $5.1\%$, and lowers PSDD by $59.3\%$.
The from-scratch sCM ablation achieves similar pointwise accuracy (R$L_2=0.282$, SSIM$=0.723$) but much worse spectral fidelity (PSDD $5.27{\times}10^{-2}$ vs.\ $1.34{\times}10^{-2}$), again indicating that teacher distillation is critical for recovering small-scale structure.
That a one-step student exceeds a 30-NFE teacher on this benchmark suggests that single-shot distillation can avoid integration error accumulation in highly turbulent fields, while inference-time OT-path noise may act as an implicit regularizer.

\begin{figure}[!h]
  \centering
  \includegraphics[width=\linewidth]{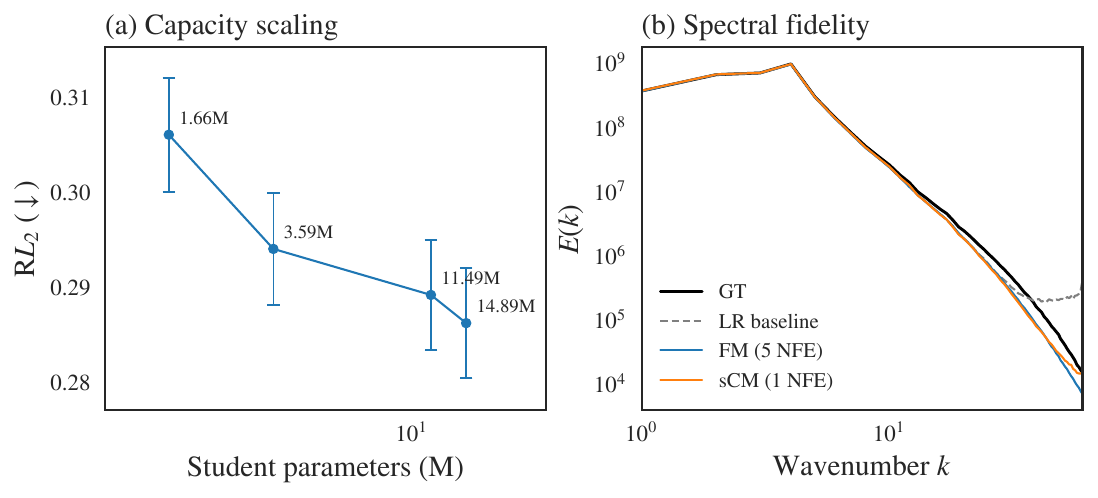}
  \caption{%
    Capacity scaling and spectral fidelity on Kolmogorov Flow ($64{\to}256$).
    \textbf{(a)} Validation R$L_2$ for four sCM students of varying
    capacity, all distilled from a common FM teacher (mean $\pm$ SEM,
    32 frames).
    \textbf{(b)} Radially-averaged power spectral density (mean, same
    32 frames) of the ground truth, nearest-neighbour LR baseline, FM teacher
    (5 NFEs), and our $14.89$\,M sCM student (1 NFE).%
  }
  \label{fig:scaling_spectrum}
\end{figure}

Figure~\ref{fig:scaling_spectrum} provides two diagnostics for the Kolmogorov student.
Panel~(a) shows that validation R$L_2$ decreases monotonically with student capacity and saturates above roughly $12$\,M parameters, where the student matches the teacher on most validation frames.
Panel~(b) reports the radially averaged power spectral density.
The nearest-neighbour LR baseline shows the expected dip near the input Nyquist frequency ($k\!\approx\!50$) and spurious high-frequency energy, whereas both the FM teacher and the 14.89\,M one-step sCM closely track the ground truth across the inertial range.
The student retains a small high-wavenumber residual, consistent with partial loss of fine-scale detail when collapsing multi-step integration into a single neural evaluation.
Spectra for the remaining benchmarks are provided in Appendix~\ref{app:qualitative}.

\begin{table}[!h]
\centering
\caption{Inference cost per frame across solvers and datasets. The distilled sCM reduces inference to a single function evaluation, yielding a substantial wall-clock speed-up over multi-step teachers at roughly half the parameter count, independent of solver, dataset, or output resolution. NFE counts model evaluations per frame: for the FM teacher, NFE\,$=$\,(outer integration steps)\,$\times$\,(stages per step), giving $5{\times}1$, $5{\times}2$, and $5{\times}6$ for Euler, Heun, and RK5 respectively. All times measured at the same per-dataset batch size (Smoke Buoyancy 32, tubulent channel flow 16, Kolmogorov Flow 8).}
\label{tab:inference_consolidated}                                                                                                      
\small                                                                                                                                  
\setlength{\tabcolsep}{6pt}
\renewcommand{\arraystretch}{1.1}                                                                                                       
\begin{tabular}{llccc}                                                                                                                  
\toprule
\textbf{Solver} & \textbf{Dataset} & \textbf{Parameters} & \textbf{NFE} & \textbf{Inference Time (s)} \\                                
\midrule                                                                                                                                
\rowcolor{gray!15} \multicolumn{5}{l}{\textit{Diffusion Models}} \\
\multirow{3}{*}{DDIM}                                                                                                                   
& Smoke Buoyancy ($128^2$)            & 3.7\,M & 30 & $0.0311 \pm 0.0001$ \\                                                                   
& Turbulent Channel Flow ($192^2$)    & 3.7\,M & 30 & $0.0719 \pm 0.0001$ \\                                                                   
& Kolmogorov Flow ($256^2$)       & 3.5\,M & 30 & $0.1319 \pm 0.0002$ \\                                                                   
\cmidrule(lr){1-5}                                                                                                                      
\multirow{3}{*}{DPM++}                                                                                                                  
& Smoke Buoyancy ($128^2$)            & 3.7\,M & 30 & $0.0314 \pm 0.0001$ \\                                                                   
& Turbulent Channel Flow ($192^2$)    & 3.7\,M & 30 & $0.0724 \pm 0.0001$ \\                                                                   
& Kolmogorov Flow ($256^2$)       & 3.5\,M & 30 & $0.1327 \pm 0.0002$ \\                                                                   
\midrule                                                                                                                                
\rowcolor{gray!15} \multicolumn{5}{l}{\textit{Teacher Models --- Flow Matching}} \\                                                     
\multirow{3}{*}{Euler}
& Smoke Buoyancy ($128^2$)            & 32.4\,M &  5 & $0.0133 \pm 0.0024$ \\
& Turbulent Channel Flow ($192^2$)    & 29.5\,M &  5 & $0.0272 \pm 0.0049$ \\
& Kolmogorov Flow ($256^2$)       & 29.5\,M &  5 & $0.0462 \pm 0.0060$ \\
\cmidrule(lr){1-5}
\multirow{3}{*}{Heun}
& Smoke Buoyancy ($128^2$)            & 32.4\,M & 10 & $0.0236 \pm 0.0032$ \\
& Turbulent Channel Flow ($192^2$)    & 29.5\,M & 10 & $0.0483 \pm 0.0053$ \\
& Kolmogorov Flow ($256^2$)       & 29.5\,M & 10 & $0.0836 \pm 0.0066$ \\
\cmidrule(lr){1-5}
\multirow{3}{*}{RK5}
& Smoke Buoyancy ($128^2$)            & 32.4\,M & 30 & $0.0667 \pm 0.0028$ \\
& Turbulent Channel Flow ($192^2$)    & 29.5\,M & 30 & $0.1382 \pm 0.0061$ \\
& Kolmogorov Flow ($256^2$)       & 29.5\,M & 30 & $0.2415 \pm 0.0006$ \\
\midrule
\rowcolor{gray!15} \multicolumn{5}{l}{\textit{\textbf{Consistency Model --- 1-step (Ours)}}} \\
\multirow{3}{*}{\textbf{sCM}}
& \textbf{Smoke Buoyancy ($128^2$)}            & \textbf{15.9\,M} & \textbf{1} & $\mathbf{0.0056 \pm 0.0021}$ \\
& \textbf{Turbulent Channel Flow ($192^2$)}    & \textbf{14.9\,M} & \textbf{1} & $\mathbf{0.0116 \pm 0.0045}$ \\
& \textbf{Kolmogorov Flow ($256^2$)}       & \textbf{14.9\,M} & \textbf{1} & $\mathbf{0.0197 \pm 0.0090}$ \\
\bottomrule
\end{tabular}
\end{table}

Table~\ref{tab:inference_consolidated} consolidates per-frame inference time across solvers and datasets.
Three trends emerge.
First, the FM teacher cost scales approximately linearly with the number of model evaluations per integration step: Heun is $\sim\!1.8{\times}$ slower than Euler, while RK5 is $\sim\!5{\times}$ slower, indicating that forward passes dominate runtime at the chosen batch sizes.
Second, the distilled sCM delivers a consistent $\sim\!12{\times}$ wall-clock speed-up over the RK5 FM teacher across all resolutions ($128^2$, $192^2$, and $256^2$), with speed-up factors of $11.9{\times}$, $11.9{\times}$, and $12.3{\times}$, respectively.
This near-constant ratio reflects the reduction from $30$ to $1$ NFE together with the student's smaller backbone.
Third, the diffusion baselines remain $5.5$--$6.7{\times}$ slower than sCM despite using a $\sim\!4{\times}$ smaller backbone, because their 30-step sampling budget is unchanged; DPM-Solver++ improves the trajectory discretization but not its length.
Overall, consistency distillation removes the dominant inference-time cost while preserving reconstruction quality, with the largest practical gains expected at high spatial resolutions where multi-step samplers are most expensive.
Qualitative super-resolution results are shown in Appendix~\ref{app:qualitative} (Figure~\ref{fig:kmflow_trajectory}).

\section{Conclusion}

We introduced one-step consistency distillation from a flow-matching teacher for fidelity enhancement in two-dimensional fluid systems.
Conditioning is supplied only at inference through OT-path trajectory initialization, allowing both teacher and student to remain unconditional models of the high-resolution distribution.

Across the Smoke Buoyancy, Turbulent Channel Flow, and Kolmogorov Flow benchmarks, the distilled student closely matches the multi-step OT-FM teacher on pointwise and spectral metrics while using roughly half the parameters and only one neural function evaluation.
This yields an approximately $12{\times}$ wall-clock speed-up at comparable reconstruction quality.
Under a matched training budget, the distilled student also improves SSIM over a from-scratch sCM by $23.1\%$ on Smoke Buoyancy, suggesting that the teacher trajectory provides an effective training curriculum rather than merely accelerating sampling.

Several limitations remain.
The fidelity--realism trade-off is controlled by a single dataset-specific noising time $\tau$; adaptive or content-dependent schedules may further reduce the residual spectral gap.
Our diffusion baselines use smaller backbones than the FM teacher and student, leaving parameter-matched comparisons to future work.
Finally, our experiments are limited to two-dimensional, statistically stationary fields, and the conditioning mechanism assumes that the upsampled low-resolution input is well approximated by an on-trajectory OT state.
Extending this framework to three-dimensional turbulence, non-stationary boundary conditions, sparse or unstructured observations, and other scientific domains such as weather and combustion is a natural direction for future work.

More broadly, our results suggest a transferable recipe: as larger flow-matching models for scientific data become available, consistency distillation can convert them into compact, deployable reconstruction models without retraining the teacher or adding task-specific conditioning machinery.


\bibliographystyle{unsrtnat}
\bibliography{ref}

\newpage
\appendix

\startcontents[appendix]
\section*{Appendix}
\printcontents[appendix]{}{1}{\setcounter{tocdepth}{2}}
\clearpage

\section{Background and Algorithms}
\label{app:background}

This appendix consolidates the technical material underlying the flow-matching teacher and the TrigFlow consistency student: the full FM objective and its conditional reformulation (Appendix~\ref{app:fm}), the TrigFlow parameterization and its lossless equivalence with the OT-FM path (Appendix~\ref{app:trigflow}), the simplified continuous-time consistency distillation loss with its JVP-based tangent estimator (Appendix~\ref{app:cm}), and the per-step training and distillation algorithms (Appendix~\ref{app:algorithms}).

\subsection{Flow Matching}
\label{app:fm}

Flow Matching constructs a continuous-time flow transforming a simple base distribution $p_{\text{base}}$ (e.g.\ a Gaussian) into a complex target data distribution $p_{\text{data}}$~\cite{lipman2022flow}. The flow is characterized by a time-indexed family of densities $\{p_t(\mathbf{z}_t)\}_{t \in [0,1]}$ together with a velocity field $\mathbf{v}_t(\mathbf{z})$ whose dynamics generate the path.

\paragraph{Conditional paths.}
A widely used construction specifies the path through a conditional coupling. Given a data sample $\mathbf{x} \sim p_{\text{data}}$ and noise $\boldsymbol{\epsilon} \sim p_{\text{base}}$, one defines a time-dependent interpolation
\begin{equation}
\mathbf{z}_t = a_t\,\mathbf{x} + b_t\,\boldsymbol{\epsilon},
\label{eq:fm_path}
\end{equation}
where $a_t, b_t : [0,1] \to \mathbb{R}$ are differentiable scalar schedules satisfying $a_0 = b_1 = 1$ and $a_1 = b_0 = 0$. Differentiating Eq.~\ref{eq:fm_path} yields the corresponding conditional velocity
\begin{equation}
\mathbf{v}(\mathbf{z}_t \mid \mathbf{x}, \boldsymbol{\epsilon}) = \frac{\mathrm{d}\mathbf{z}_t}{\mathrm{d}t} = a_t'\,\mathbf{x} + b_t'\,\boldsymbol{\epsilon}.
\end{equation}

\paragraph{Flow matching objective.}
Let $\mathbf{v}_\phi(\mathbf{z}, t)$ denote the parametrized teacher velocity model. Ideally, one would match $\mathbf{v}_\phi$ to the \emph{marginal} velocity field that generates the density path $\{p_t\}$,
\begin{equation}
\mathcal{L}_{\text{FM}} = \mathbb{E}_{t,\, \mathbf{z}_t \sim p_t} \Bigl\| \mathbf{v}_\phi(\mathbf{z}_t, t) - \mathbb{E}_{p_t(\mathbf{x}, \boldsymbol{\epsilon} \mid \mathbf{z}_t)}\!\bigl[\mathbf{v}(\mathbf{z}_t \mid \mathbf{x}, \boldsymbol{\epsilon})\bigr] \Bigr\|_2^2.
\end{equation}
The inner conditional expectation is generally intractable, as it requires posterior inference under the coupling. Conditional FM (CFM)~\cite{lipman2022flow} avoids this difficulty by directly matching the \emph{conditional} velocity and leveraging the fact that its gradient coincides with that of the ideal FM objective,
\begin{equation}
\mathcal{L}_{\text{CFM}} = \mathbb{E}_{t,\, \mathbf{x} \sim p_{\text{data}},\, \boldsymbol{\epsilon} \sim p_{\text{base}}} \Bigl\| \mathbf{v}_\phi(\mathbf{z}_t, t) - \mathbf{v}(\mathbf{z}_t \mid \mathbf{x}, \boldsymbol{\epsilon}) \Bigr\|_2^2.
\label{eq:cfm}
\end{equation}

\paragraph{Optimal-transport path.}
The linear schedule $a_t = 1 - t$, $b_t = t$ used in the main paper (Eq.~\ref{eq:ot_path}) yields the interpolation $\mathbf{z}_t = (1 - t)\,\mathbf{x} + t\,\boldsymbol{\epsilon}$ and a constant conditional velocity $\mathbf{v}(\mathbf{z}_t \mid \mathbf{x}, \boldsymbol{\epsilon}) = \boldsymbol{\epsilon} - \mathbf{x}$. This corresponds to straight-line trajectories in latent space and is referred to as the optimal-transport (OT) path~\cite{lipman2022flow}. Our teacher is trained under this parameterization.

\subsection{TrigFlow Parameterization and OT Equivalence}
\label{app:trigflow}

\paragraph{TrigFlow.}
TrigFlow~\cite{lu2024simplifying} replaces the linear OT schedule with a sinusoidal coupling
\begin{equation}
\mathbf{z}_t = \cos(t)\,\mathbf{x} + \sin(t)\,\boldsymbol{\epsilon}, \qquad t \in [0, \tfrac{\pi}{2}],
\label{eq:trigflow_path_app}
\end{equation}
with conditional velocity $\mathbf{v}_t = \cos(t)\,\boldsymbol{\epsilon} - \sin(t)\,\mathbf{x}$. The continuous-time consistency framework of~\cite{lu2024simplifying} is derived under this parameterization.

\paragraph{OT--TrigFlow equivalence.}
\citet{chen2025sanasprint} establish a lossless equivalence between OT-FM and TrigFlow. Let $t_{\text{Trig}} \in [0, \pi/2]$ and $t_{\text{OT}} \in [0, 1]$. The transforms are:
\begin{align}
t_{\text{OT}} &= \frac{\sin(t_{\text{Trig}})}{\sin(t_{\text{Trig}}) + \cos(t_{\text{Trig}})}, \label{eq:time_transform_app} \\[4pt]
\mathbf{z}_{t,\text{OT}} &= \frac{\mathbf{z}_{t,\text{Trig}}}{\sqrt{t_{\text{OT}}^2 + (1 - t_{\text{OT}})^2}}, \label{eq:latent_transform_app} \\[4pt]
\mathbf{F}_\theta(\mathbf{z}_{t,\text{Trig}}, t_{\text{Trig}}) &= \frac{(1 - 2t_{\text{OT}})\,\mathbf{z}_{t,\text{OT}} + \bigl(1 - 2t_{\text{OT}} + 2t_{\text{OT}}^2\bigr)\,\mathbf{v}_{\text{OT}}}{\sqrt{t_{\text{OT}}^2 + (1 - t_{\text{OT}})^2}}, \label{eq:velocity_transform_app}
\end{align}
where $\mathbf{v}_{\text{OT}}$ is the OT-FM velocity at $(\mathbf{z}_{t,\text{OT}}, t_{\text{OT}})$. Equation~\ref{eq:time_transform_app} maps TrigFlow time onto the OT interval, Eq.~\ref{eq:latent_transform_app} rescales the latent for the differing path geometries, and Eq.~\ref{eq:velocity_transform_app} converts the OT velocity into its TrigFlow counterpart. Together these transforms allow our pretrained OT-FM teacher to provide TrigFlow-consistent supervision to the consistency student during distillation.

\subsection{Consistency Models and sCD Loss}
\label{app:cm}

\paragraph{TrigFlow parameterization of the consistency function.}
Working in TrigFlow coordinates, the consistency function is parameterized in terms of the underlying velocity network as
\begin{equation}
f_\theta(\mathbf{z}_{t,\text{Trig}}, t_{\text{Trig}}) = \cos(t_{\text{Trig}})\,\mathbf{z}_{t,\text{Trig}} - \sin(t_{\text{Trig}})\,\mathbf{F}_\theta(\mathbf{z}_{t,\text{Trig}}, t_{\text{Trig}}),
\label{eq:f_theta_app}
\end{equation}
which automatically enforces the boundary condition $f_\theta(\mathbf{z}_0, 0) = \mathbf{x}$.

\paragraph{Simplified continuous-time consistency distillation (sCD).}
Discrete-time consistency models train the self-consistency identity by sampling pairs $(t, t + \Delta t)$, but the resulting finite-difference approximation introduces discretization error~\cite{lu2024simplifying}. The simplified continuous-time framework differentiates the self-consistency loss directly. The sCD loss is
\begin{equation}
\mathcal{L}_{\text{sCD}}(\theta) = \mathbb{E}_{\mathbf{z}_t,\, t} \biggl\| \mathbf{F}_\theta(\mathbf{z}_t, t) - \mathbf{F}_{\theta^-}(\mathbf{z}_t, t) - \cos(t)\,\frac{\mathrm{d}}{\mathrm{d}t} f_{\theta^-}(\mathbf{z}_t, t) \biggr\|_2^2,
\label{eq:scd_loss_app}
\end{equation}
where $\theta^-$ is an EMA of the student parameters with stop-gradient. The EMA target stabilizes the self-supervised objective by decoupling the regression target from the parameters being optimized.

\paragraph{JVP-based tangent estimation.}
The tangent $\frac{\mathrm{d}}{\mathrm{d}t} f_{\theta^-}$ in Eq.~\ref{eq:scd_loss_app} decomposes as
\begin{equation}
\frac{\mathrm{d}}{\mathrm{d}t} f_{\theta^-}(\mathbf{z}_t, t) = \frac{\partial f_{\theta^-}}{\partial \mathbf{z}_t} \cdot \frac{\mathrm{d}\mathbf{z}_t}{\mathrm{d}t} + \frac{\partial f_{\theta^-}}{\partial t},
\end{equation}
and is computed exactly via a single Jacobian-vector product (JVP) against the tangent vector $(\mathrm{d}\mathbf{z}_t/\mathrm{d}t,\, 1)$, where $\mathrm{d}\mathbf{z}_t/\mathrm{d}t$ is supplied by the teacher~\cite{lu2024simplifying}. This avoids storing or differencing the network across multiple timesteps.

\subsection{Algorithms}
\label{app:algorithms}

The two algorithms below give the per-step training procedures for the two stages of our pipeline. Algorithm~\ref{alg:fm_training} is the standard conditional flow-matching update (Eq.~\ref{eq:cfm}) instantiated under the linear OT path of Eq.~\ref{eq:ot_path}, and is used to train the unconditional OT-FM teacher on the high-resolution distribution. Algorithm~\ref{alg:scm_distillation_loss} then distils that frozen teacher into a one-step TrigFlow consistency student: it samples a TrigFlow timestep, evaluates the teacher in OT coordinates via the lossless transforms of Appendix~\ref{app:trigflow}, and uses the resulting trajectory tangent inside a single Jacobian--vector product to construct the sCD target of Eq.~\ref{eq:scd_loss_app}. The EMA copy $\mathbf{F}_{\theta^-}$ is updated after each optimizer step to provide a stop-gradient regression target on the next iteration.

\begin{algorithm}[H]
\caption{Single training step for the OT-FM teacher}
\label{alg:fm_training}
\begin{algorithmic}[1]
\REQUIRE High-resolution sample $\mathbf{x}$, velocity field $\mathbf{v}_\phi$, time sampler $p(t)$ on $(0,1)$
\STATE Sample time $t \sim p(t)$, noise $\boldsymbol{\epsilon} \sim \mathcal{N}(\mathbf{0}, \mathbf{I})$
\STATE $\mathbf{z}_t \leftarrow (1 - t)\,\mathbf{x} + t\,\boldsymbol{\epsilon}$
\STATE $\mathbf{v}_t \leftarrow \boldsymbol{\epsilon} - \mathbf{x}$
\STATE $\hat{\mathbf{v}} \leftarrow \mathbf{v}_\phi(\mathbf{z}_t, t)$
\STATE $\ell \leftarrow \|\hat{\mathbf{v}} - \mathbf{v}_t\|_2^2$
\STATE Update $\phi$ with one optimizer step using $\nabla_\phi \ell$
\end{algorithmic}
\end{algorithm}

\begin{algorithm}[H]
\caption{Single distillation step (OT-FM teacher $\rightarrow$ TrigFlow sCM student)}
\label{alg:scm_distillation_loss}
\begin{algorithmic}[1]
\REQUIRE Student $\mathbf{F}_\theta$, EMA student $\mathbf{F}_{\theta^-}$, OT-FM teacher $\mathbf{v}_\phi$, clean sample $\mathbf{x}$, TrigFlow--OT transforms $\mathcal{T}_t, \mathcal{T}_z, \mathcal{T}_v$ (Eqs.~\ref{eq:time_transform_app}--\ref{eq:velocity_transform_app}), TrigFlow time sampler $p_{\text{Trig}}$
\STATE $t_{\text{Trig}} \sim p_{\text{Trig}}$, $\boldsymbol{\epsilon} \sim \mathcal{N}(\mathbf{0}, \mathbf{I})$
\STATE $\mathbf{z}_{t,\text{Trig}} \leftarrow \mathbf{x} \cos(t_{\text{Trig}}) + \boldsymbol{\epsilon} \sin(t_{\text{Trig}})$
\STATE $t_{\text{OT}} \leftarrow \mathcal{T}_t(t_{\text{Trig}})$, $\mathbf{z}_{t,\text{OT}} \leftarrow \mathcal{T}_z(\mathbf{z}_{t,\text{Trig}}, t_{\text{OT}})$
\STATE $\mathbf{v}_{\text{OT}} \leftarrow \mathbf{v}_\phi(\mathbf{z}_{t,\text{OT}}, t_{\text{OT}})$
\STATE $\hat{\mathbf{F}}_{\text{teacher}} \leftarrow \mathcal{T}_v(\mathbf{z}_{t,\text{OT}}, t_{\text{OT}}, \mathbf{v}_{\text{OT}})$, $\mathbf{F}_{\text{stud}} \leftarrow \mathbf{F}_\theta(\mathbf{z}_{t,\text{Trig}}, t_{\text{Trig}})$, $\mathbf{F}_{\text{ema}} \leftarrow \mathbf{F}_{\theta^-}(\mathbf{z}_{t,\text{Trig}}, t_{\text{Trig}})$
\STATE $f_{\theta^-}(\mathbf{z}_{t,\text{Trig}}, t_{\text{Trig}}) \leftarrow \cos(t_{\text{Trig}})\,\mathbf{z}_{t,\text{Trig}} - \sin(t_{\text{Trig}})\,\mathbf{F}_{\text{ema}}$
\STATE $\frac{\mathrm{d}}{\mathrm{d}t} f_{\theta^-} \leftarrow \mathrm{JVP}\bigl(f_{\theta^-},\,(\mathbf{z}_{t,\text{Trig}}, t_{\text{Trig}}),\,(\hat{\mathbf{F}}_{\text{teacher}}, \mathbf{1})\bigr)$
\STATE $\text{target} \leftarrow \mathrm{stopgrad}\bigl(\mathbf{F}_{\text{ema}} + \cos(t_{\text{Trig}})\,\tfrac{\mathrm{d}}{\mathrm{d}t} f_{\theta^-}\bigr)$
\STATE $\mathcal{L}_{\text{sCD}} \leftarrow \|\mathbf{F}_{\text{stud}} - \text{target}\|_2^2$
\STATE Update $\theta$ with one optimizer step using $\nabla_\theta \mathcal{L}_{\text{sCD}}$
\end{algorithmic}
\end{algorithm}

\clearpage

\section{Baseline Solver Details}
\label{app:solvers}

This appendix specifies the multi-step solvers used by the diffusion and flow-matching baselines reported in Section~\ref{sec:experiments} (Tables~\ref{tab:smoke_metrics}--\ref{tab:kf_metrics} and~\ref{tab:inference_consolidated}). All three solvers share the inference-time conditioning of Section~\ref{sec:conditioning}, with the starting state $\mathbf{x}_{\tau_0}$ formed by noise injection (Eq.~\ref{eq:noise_injection}) at the per-dataset $\tau$.

\subsection{Denoising Diffusion Implicit Models}
\label{app:ddim}

Denoising Diffusion Implicit Models (DDIM)~\cite{song2022denoisingdiffusionimplicitmodels} accelerates DDPM~\cite{ho2020denoising} sampling by replacing the Markovian reverse chain with a deterministic non-Markovian update that preserves the marginals of the forward diffusion process, so that fewer steps suffice while a consistent trajectory is preserved. Given a learned noise predictor $\epsilon_\theta(\mathbf{x}_t, t)$ and a noise schedule $\{\bar\alpha_t\}_{t=0}^{T}$, the deterministic step from time $s$ to time $u<s$ uses the implied clean prediction
\begin{equation}
\hat{\mathbf{x}}_0(\mathbf{x}_s, s) \;=\; \frac{\mathbf{x}_s - \sqrt{1-\bar\alpha_s}\,\epsilon_\theta(\mathbf{x}_s, s)}{\sqrt{\bar\alpha_s}},
\end{equation}
to advance the sample as
\begin{equation}
\mathbf{x}_u \;=\; \sqrt{\bar\alpha_u}\,\hat{\mathbf{x}}_0(\mathbf{x}_s, s) \;+\; \sqrt{1-\bar\alpha_u}\,\epsilon_\theta(\mathbf{x}_s, s).
\end{equation}
We use 30 sampling steps on a uniformly-spaced grid over $[0, T]$. The full procedure is summarised in Algorithm~\ref{alg:ddim}.

\begin{algorithm}[H]
\caption{DDIM sampling}
\label{alg:ddim}
\begin{algorithmic}[1]
\REQUIRE Noise predictor $\epsilon_\theta$, schedule $\{\bar\alpha_t\}$, step grid $T=\tau_0>\tau_1>\cdots>\tau_K=0$ (we use $K=30$), starting state $\mathbf{x}_{\tau_0}$
\FOR{$k=0,\dots,K-1$}
  \STATE $s \gets \tau_k$,\; $u \gets \tau_{k+1}$
  \STATE $\boldsymbol{\epsilon}_s \gets \epsilon_\theta(\mathbf{x}_s, s)$
  \STATE $\hat{\mathbf{x}}_0 \gets \bigl(\mathbf{x}_s - \sqrt{1-\bar\alpha_s}\,\boldsymbol{\epsilon}_s\bigr)\big/\sqrt{\bar\alpha_s}$
  \STATE $\mathbf{x}_u \gets \sqrt{\bar\alpha_u}\,\hat{\mathbf{x}}_0 + \sqrt{1-\bar\alpha_u}\,\boldsymbol{\epsilon}_s$
\ENDFOR
\end{algorithmic}
\textbf{Return:} High-resolution sample $\mathbf{x}_0$.
\end{algorithm}

\subsection{DPM-Solver++}
\label{app:dpmpp}

DPM-Solver++~\cite{Lu_2025} reformulates the reverse diffusion ODE in log-SNR coordinates and integrates it with a multistep method, exploiting the semi-linear structure of the diffusion ODE to attain higher effective order than DDIM at the same number of function evaluations. With $\lambda_t = \log\bigl(\sqrt{\bar\alpha_t}/\sqrt{1-\bar\alpha_t}\bigr)$ as the log-SNR variable, $\sigma_t = \sqrt{1-\bar\alpha_t}$, and $\hat{\mathbf{x}}_0(\mathbf{x}_t, t)$ as defined for DDIM in Appendix~\ref{app:ddim}, the second-order multistep variant DPM-Solver++(2M), which we use here, advances the sample from $\lambda_s$ to $\lambda_u$ via
\begin{equation}
\mathbf{x}_u \;=\; \frac{\sigma_u}{\sigma_s}\,\mathbf{x}_s \;-\; \sqrt{\bar\alpha_u}\,\bigl(e^{-h_k}-1\bigr)\,\mathbf{D}_k,
\qquad
h_k = \lambda_u - \lambda_s,
\end{equation}
where the multistep operator $\mathbf{D}_k$ blends the current and previous clean predictions according to
\begin{equation}
\mathbf{D}_k \;=\; \Bigl(1+\tfrac{1}{2 r_k}\Bigr)\,\hat{\mathbf{x}}_0(\mathbf{x}_s, s) \;-\; \tfrac{1}{2 r_k}\,\hat{\mathbf{x}}_0(\mathbf{x}_{s_{k-1}}, s_{k-1}),
\qquad
r_k = h_{k-1}/h_k,
\end{equation}
with $\mathbf{D}_0 = \hat{\mathbf{x}}_0(\mathbf{x}_{\tau_0}, \tau_0)$ for the first step. We use 30 steps on the same uniformly-spaced grid as DDIM. The full procedure is summarised in Algorithm~\ref{alg:dpmpp}.

\begin{algorithm}[H]
\caption{DPM-Solver++(2M) sampling}
\label{alg:dpmpp}
\begin{algorithmic}[1]
\REQUIRE Noise predictor $\epsilon_\theta$, schedule $\{\bar\alpha_t\}$, step grid $T=\tau_0>\tau_1>\cdots>\tau_K=0$ (we use $K=30$), starting state $\mathbf{x}_{\tau_0}$
\STATE Pre-compute $\lambda_k = \log\bigl(\sqrt{\bar\alpha_{\tau_k}}/\sqrt{1-\bar\alpha_{\tau_k}}\bigr)$ and $\sigma_k = \sqrt{1-\bar\alpha_{\tau_k}}$ for $k=0,\dots,K$
\STATE $\hat{\mathbf{x}}_0^{\,(0)} \gets \bigl(\mathbf{x}_{\tau_0} - \sigma_0\,\epsilon_\theta(\mathbf{x}_{\tau_0}, \tau_0)\bigr)\big/\sqrt{\bar\alpha_{\tau_0}}$
\FOR{$k=0,\dots,K-1$}
  \STATE $h_k \gets \lambda_{k+1} - \lambda_k$
  \IF{$k=0$}
    \STATE $\mathbf{D}_k \gets \hat{\mathbf{x}}_0^{\,(k)}$
  \ELSE
    \STATE $r_k \gets h_{k-1}/h_k$
    \STATE $\mathbf{D}_k \gets \bigl(1+1/(2 r_k)\bigr)\,\hat{\mathbf{x}}_0^{\,(k)} - \bigl(1/(2 r_k)\bigr)\,\hat{\mathbf{x}}_0^{\,(k-1)}$
  \ENDIF
  \STATE $\mathbf{x}_{\tau_{k+1}} \gets (\sigma_{k+1}/\sigma_k)\,\mathbf{x}_{\tau_k} - \sqrt{\bar\alpha_{\tau_{k+1}}}\,\bigl(e^{-h_k}-1\bigr)\,\mathbf{D}_k$
  \STATE $\hat{\mathbf{x}}_0^{\,(k+1)} \gets \bigl(\mathbf{x}_{\tau_{k+1}} - \sigma_{k+1}\,\epsilon_\theta(\mathbf{x}_{\tau_{k+1}}, \tau_{k+1})\bigr)\big/\sqrt{\bar\alpha_{\tau_{k+1}}}$
\ENDFOR
\end{algorithmic}
\textbf{Return:} High-resolution sample $\mathbf{x}_{\tau_K}$.
\end{algorithm}

\subsection{Heun Method}
\label{app:heun}

The Heun (improved-Euler) method is a two-stage explicit Runge--Kutta integrator that is second-order accurate in the step size~\cite{hairer1993solving}. Applied to the unconditional flow-matching ODE $\mathrm{d}\mathbf{z}/\mathrm{d}t = \mathbf{v}_\phi(\mathbf{z}, t)$, the step from time $t$ to time $t' = t + \Delta t$ uses one predictor evaluation followed by one corrector evaluation,
\begin{equation}
\mathbf{k}_1 = \mathbf{v}_\phi(\mathbf{z}_t, t),
\quad
\tilde{\mathbf{z}} = \mathbf{z}_t + \Delta t\,\mathbf{k}_1,
\quad
\mathbf{k}_2 = \mathbf{v}_\phi(\tilde{\mathbf{z}}, t'),
\quad
\mathbf{z}_{t'} = \mathbf{z}_t + \tfrac{\Delta t}{2}\bigl(\mathbf{k}_1 + \mathbf{k}_2\bigr),
\end{equation}
costing 2 NFEs per outer step. We integrate backward from $t = \tau$ to $t = 0$ over 5 outer steps (10 NFEs total) on a uniform grid. The full procedure is summarised in Algorithm~\ref{alg:heun}.

\begin{algorithm}[H]
\caption{Heun sampling for the flow-matching ODE}
\label{alg:heun}
\begin{algorithmic}[1]
\REQUIRE Velocity field $\mathbf{v}_\phi$, step grid $\tau = t_0 > t_1 > \cdots > t_K = 0$ (we use $K = 5$), starting state $\mathbf{z}_{t_0}$
\FOR{$k=0,\dots,K-1$}
  \STATE $\Delta t \gets t_{k+1} - t_k$
  \STATE $\mathbf{k}_1 \gets \mathbf{v}_\phi(\mathbf{z}_{t_k}, t_k)$
  \STATE $\tilde{\mathbf{z}} \gets \mathbf{z}_{t_k} + \Delta t\,\mathbf{k}_1$
  \STATE $\mathbf{k}_2 \gets \mathbf{v}_\phi(\tilde{\mathbf{z}}, t_{k+1})$
  \STATE $\mathbf{z}_{t_{k+1}} \gets \mathbf{z}_{t_k} + \tfrac{\Delta t}{2}\,\bigl(\mathbf{k}_1 + \mathbf{k}_2\bigr)$
\ENDFOR
\end{algorithmic}
\textbf{Return:} High-resolution sample $\mathbf{z}_{t_K}$.
\end{algorithm}

\clearpage

\section{Per-Dataset Qualitative Samples and Power Spectra}
\label{app:qualitative}

This appendix collects the per-dataset super-resolved comparisons and power spectra that are not shown in the main paper. The qualitative figures follow the same layout as Figure~\ref{fig:smoke_trajectory} --- row 1 is the nearest-neighbour-upsampled low-resolution input, row 2 is the high-resolution ground truth, row 3 is the one-step sCM student reconstruction, and rows 4--5 are the per-pixel absolute residuals of the sCM student and the multi-step OT-FM teacher against the ground truth. The three field rows share a single field colorbar and the two error rows share a single sequential error colorbar, so brighter pixels denote larger absolute deviation from the ground truth.

\begin{figure}[!h]
  \centering
  \includegraphics[width=\linewidth]{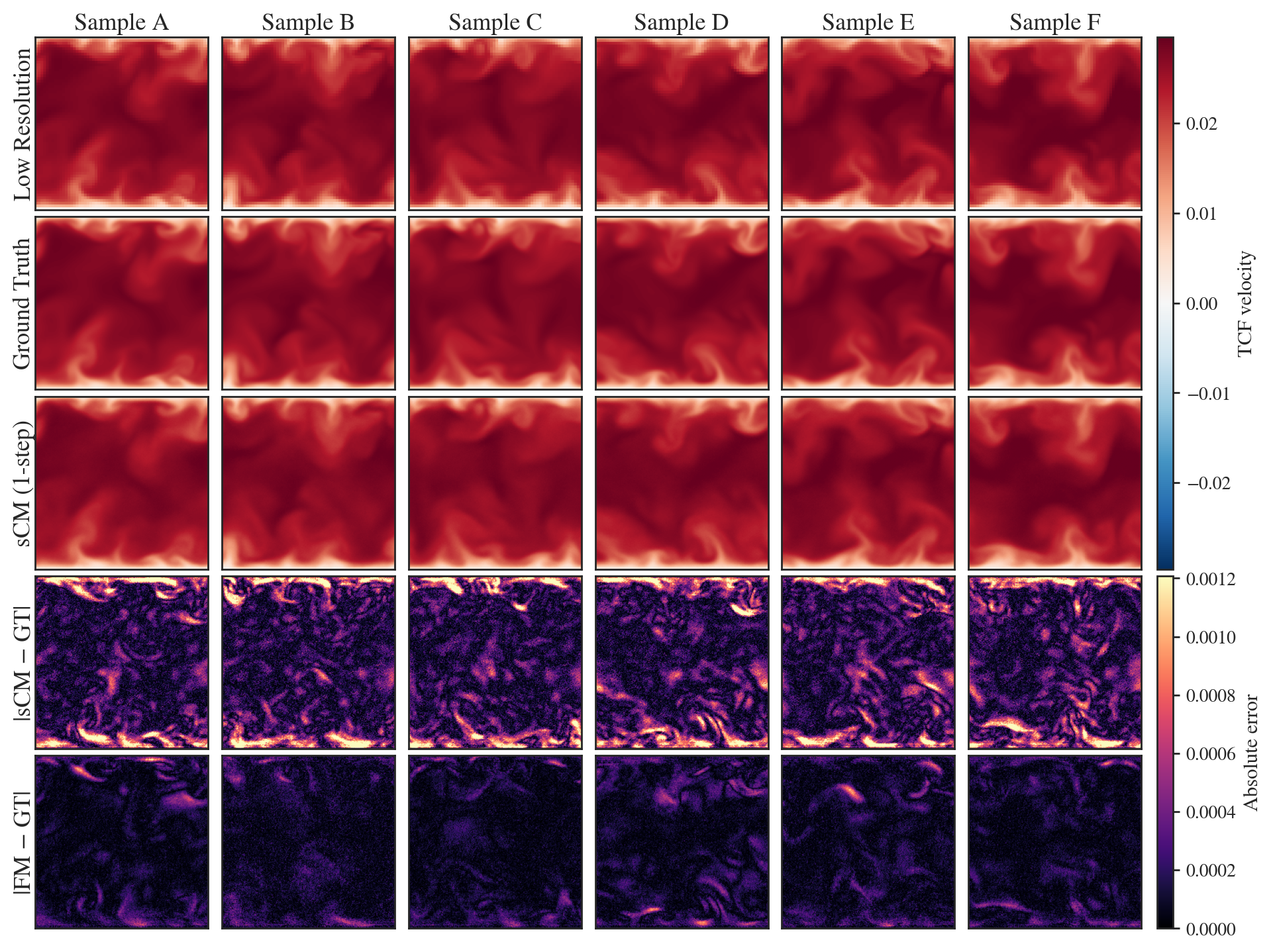}
  \caption{\textbf{Turbulent Channel Flow super-resolved samples} ($64{\to}192$). Six independently super-resolved frames (Samples A--F) drawn from across the dataset; columns are i.i.d.\ reconstructions. Layout matches Figure~\ref{fig:smoke_trajectory}. The streamwise-velocity field is dominated by the bulk channel profile, so visual differences between the nearest-neighbour-upsampled input, ground truth, and sCM reconstruction are subtle. The error rows make the residual gap explicit: the multi-step OT-FM teacher achieves roughly a third of the sCM student's pointwise error here, consistent with the metric ratios in Table~\ref{tab:tcf_metrics}. The student residual is concentrated in the near-wall streaks where the small-scale fluctuations live; the bulk profile is recovered to within machine-precision-scale error in both models.}
  \label{fig:tcf_trajectory}
\end{figure}

\begin{figure}[!h]
  \centering
  \includegraphics[width=\linewidth]{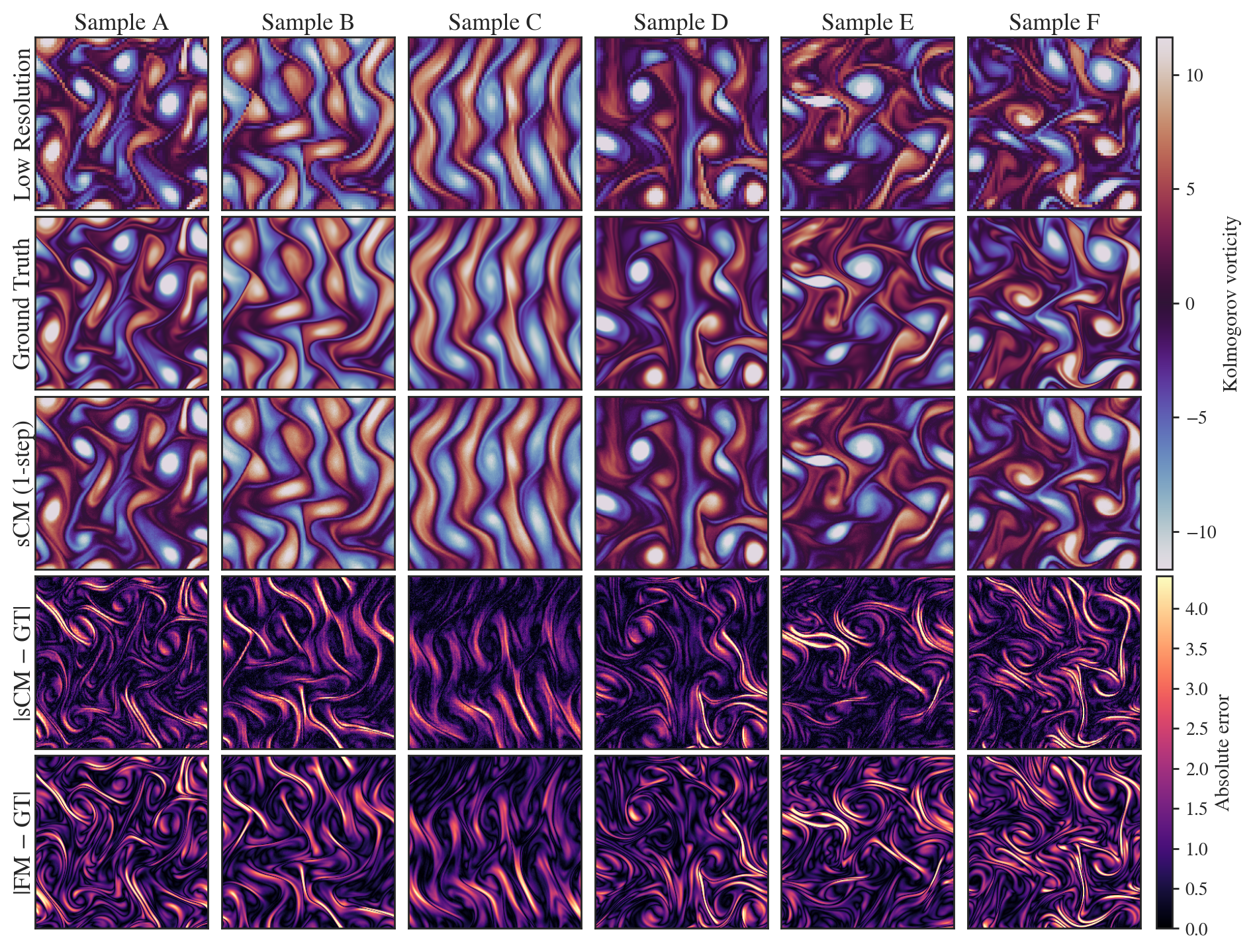}
  \caption{\textbf{Kolmogorov Flow super-resolved samples} ($64{\to}256$). Six independently super-resolved frames (Samples A--F) drawn from across the dataset; columns are i.i.d.\ reconstructions. Layout matches Figure~\ref{fig:smoke_trajectory}. The vorticity field exhibits dense multi-scale structure with prominent same-sign vortex patches and steep shear sheets between them. The one-step sCM student recovers the location, sign, and amplitude of the vortices, and the residual rows show that its absolute error is comparable to and, on every sample shown here, modestly lower than the multi-step OT-FM teacher's, consistent with the metric ratios in Table~\ref{tab:kf_metrics}.}
  \label{fig:kmflow_trajectory}
\end{figure}

\paragraph{Reproducibility.} Samples A--F of Figure~\ref{fig:smoke_trajectory} (Smoke Buoyancy) are at indices $4173, 30000, 50000, 78901, 135000, 165000$ of \texttt{128\_smoke\_hr.npy}. Samples A--F of Figure~\ref{fig:tcf_trajectory} (TCF) are at indices $1287, 9461, 14906, 24561, 29871, 35402$ of \texttt{192\_tcf\_hr.npy}. Samples A--F of Figure~\ref{fig:kmflow_trajectory} (Kolmogorov Flow) are at indices $487, 2589, 4823, 7184, 9523, 11947$ of \texttt{256\_kmflow.npy}. Indices were chosen to span the full dataset with non-consecutive spacing.

\begin{figure}[!h]
  \centering
  \begin{minipage}[t]{0.48\linewidth}
    \centering
    \includegraphics[width=\linewidth]{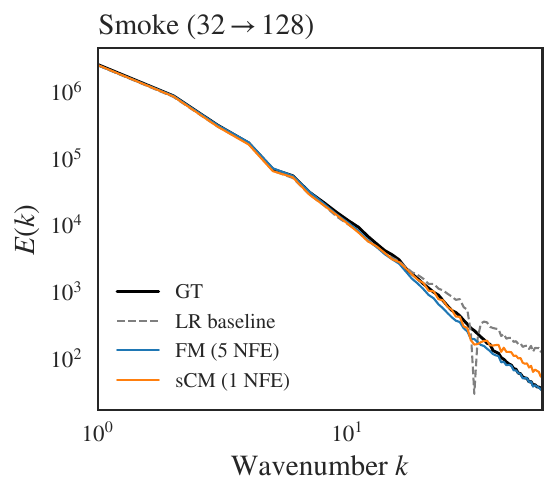}\\
    \textbf{(a)} Smoke Buoyancy ($32{\to}128$)
  \end{minipage}\hfill
  \begin{minipage}[t]{0.48\linewidth}
    \centering
    \includegraphics[width=\linewidth]{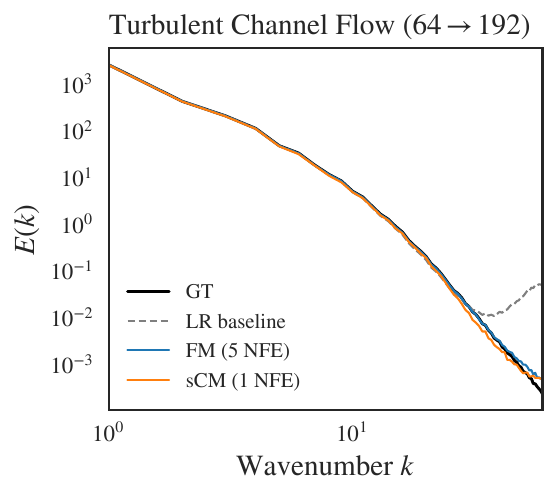}\\
    \textbf{(b)} Turbulent Channel Flow ($64{\to}192$)
  \end{minipage}
  \caption{\textbf{Power spectra on the remaining benchmarks.} Radially-averaged spectra of the ground truth, nearest-neighbour LR baseline, multi-step FM teacher, and the one-NFE sCM student ($15.9$\,M for Smoke Buoyancy, $14.9$\,M for TCF).}
  \label{fig:psd_other}
\end{figure}

\paragraph{Power spectra on the remaining benchmarks.}
We complement the Kolmogorov spectrum in Figure~\ref{fig:scaling_spectrum}(b) with the corresponding spectra on the Smoke Buoyancy and Turbulent Channel Flow benchmarks (Figure~\ref{fig:psd_other}). The protocol matches that figure --- per-channel mean-centered 2D DFT, fftshift, normalised power, and a radial average over wavenumber bins (Appendix~\ref{app:psd}); spectra are averaged over the same validation subset used for the metric tables. On Smoke Buoyancy (panel~a), the nearest-neighbour baseline shows a sharp drop near the input Nyquist with spurious energy beyond, while both the multi-step FM teacher and the one-NFE sCM student recover the ground-truth tail across the resolved range, consistent with the PSDD ranking in Table~\ref{tab:smoke_metrics}. On TCF (panel~b), the bulk channel profile dominates the low-wavenumber content, and the residual gap between the sCM student and the FM teacher at intermediate wavenumbers tracks the per-table PSDD discrepancy in Table~\ref{tab:tcf_metrics}.

\clearpage

\section{Inference-Time Sensitivity Studies}
\label{app:inference_sensitivity}

This appendix collects sensitivity studies on the two stochastic ingredients of the inference pipeline of Section~\ref{sec:conditioning}: the noise-injection time $\tau$ that governs the fidelity--realism trade-off, and the Gaussian noise vector $\boldsymbol{\epsilon}$ drawn each time a sample is generated. The first study justifies the per-dataset $\tau$ used in the main paper; the second quantifies how much of the metric values reported in Section~\ref{sec:experiments} can be attributed to the random draw of $\boldsymbol{\epsilon}$.

\subsection{Noise-strength sweep ($\tau$)}
\label{app:tau_sweep}

We sweep $\tau \in (0, 1]$ on a 50-point uniform grid and, at each grid point, run the distilled sCM student on $N_{\text{samples}} = 30$ held-out low-resolution frames, recording the RMSE against the matched ground-truth high-resolution targets. The sweep is implemented as an Optuna~\cite{akiba2019optuna} \texttt{GridSampler} study with a sqlite backend to enable resumption. Each trial draws a fresh $\boldsymbol{\epsilon}$, so the per-trial RMSE inherits a small amount of sampling noise that is characterised separately in Appendix~\ref{app:noise_sensitivity}.

\begin{figure}[!h]
  \centering
  \includegraphics[width=0.85\linewidth]{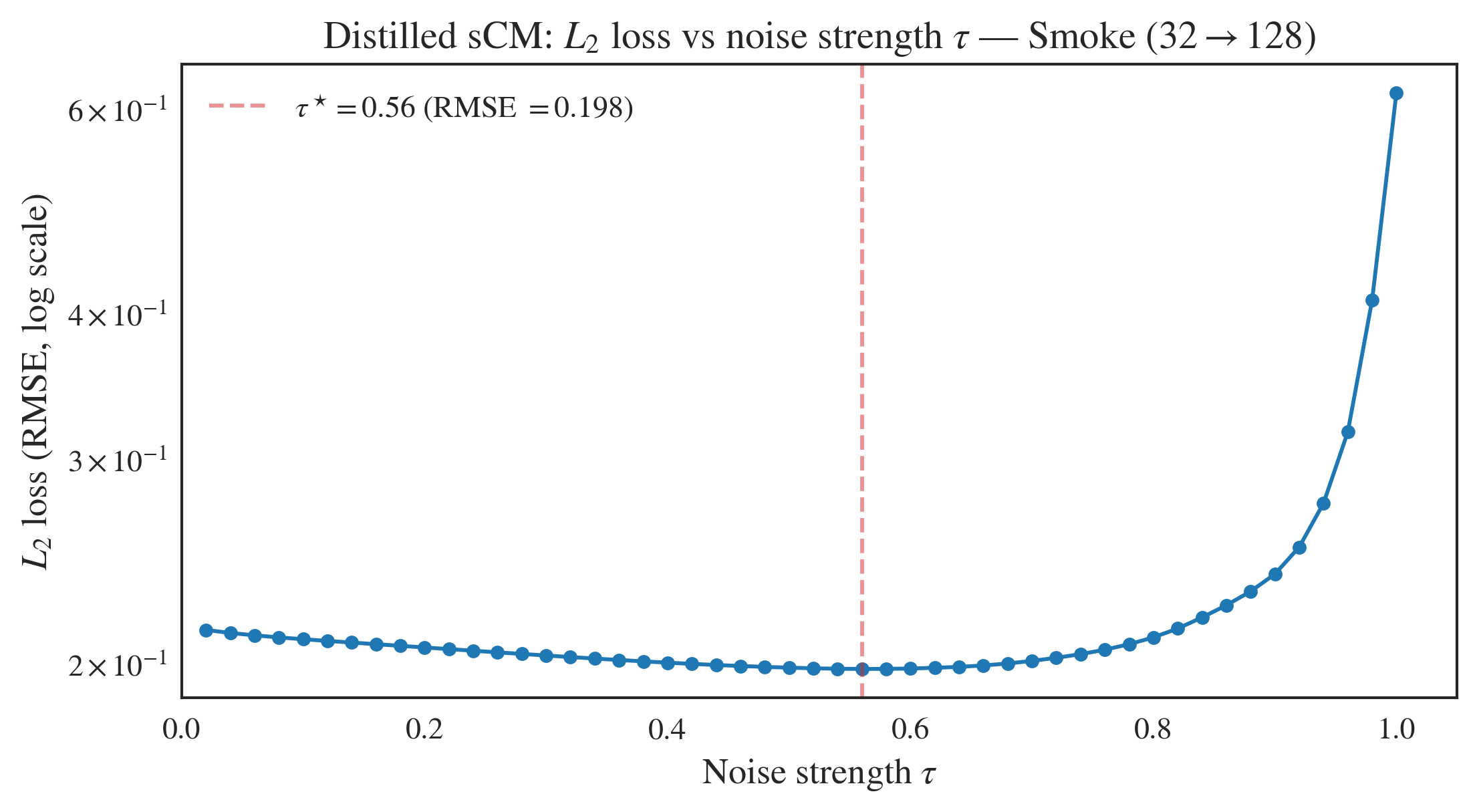}
  \caption{\textbf{Smoke Buoyancy noise-strength sweep.} RMSE of the distilled sCM student against the matched ground truth as a function of the inference noise-injection time $\tau$, on the Smoke Buoyancy ($32{\to}128$) task ($N_{\text{samples}}=30$, 50-point uniform $\tau$ grid). The empirical optimum is $\tau^{\star} = 0.56$ with RMSE $0.1977$; the default $\tau = 0.3$ used in the main paper yields RMSE $0.2031$ ($+2.7\%$ relative to the optimum), well inside the plateau (RMSE within $+5\%$ of optimum for $\tau \in [0.18, 0.76]$). RMSE rises sharply only past $\tau \approx 0.85$, reaching $0.619$ at $\tau = 1$ where the LR observation is essentially erased.}
  \label{fig:tau_sweep_smoke}
\end{figure}

The curve has the characteristic shape predicted by the trade-off in Section~\ref{sec:conditioning}: a wide low-RMSE plateau spanning roughly half the $\tau$ range, and a sharp rise as $\tau \to 1$ where the noised initialization $\mathbf{z}_\tau$ approaches pure Gaussian noise and the conditioning information from $\mathbf{x}_{\text{LR}}^{\uparrow}$ is destroyed. Within the plateau the choice of $\tau$ is largely cosmetic; the practical sensitivity emerges only near the upper boundary. The operating point $\tau = 0.3$ used throughout the main paper is suboptimal by under three percent in RMSE and lies comfortably inside the plateau, while also matching the regime in which the consistency student was trained.

\subsection{Impact of generation randomness}
\label{app:noise_sensitivity}

The sCM student is a stochastic generator: each call draws a fresh Gaussian noise vector $\boldsymbol{\epsilon}$ to form the noised initialization $\mathbf{z}_\tau$ in Eq.~\ref{eq:noise_injection}. To verify that the headline metrics reported in Section~\ref{sec:experiments} are not artefacts of a particular noise draw, we repeat the validation-split evaluation $K = 100$ times with independent random seeds, holding $\tau = 0.3$ and the model weights fixed at their main-paper values. To make each per-seed estimate itself stable, every run scores $64$ evenly-spaced frames from the held-out validation split (a 4$\times$ wider subset than the main-table evaluation), so the spread we report below is dominated by the random draw of $\boldsymbol{\epsilon}$ rather than by frame sub-sampling. For each run we record R$L_2$, SSIM, and PSDD using the implementations from Section~\ref{sec:experiments} and Appendix~\ref{appendix:metrics}.

\begin{table}[!h]
\centering
\caption{Sensitivity of the distilled sCM student's evaluation metrics on Smoke Buoyancy ($32{\to}128$) to the inference-time noise draw, over $K = 100$ random seeds at fixed $\tau = 0.3$ and a fixed $64$-frame validation subset. Reported values are mean $\pm$ standard deviation, with 5th/95th percentiles and full min/max ranges. The relative spread $\sigma / \mu$ characterises how much a single-seed evaluation can shift in either direction.}
\label{tab:noise_sensitivity_smoke}
\setlength{\tabcolsep}{3pt}
\begin{tabular}{lcccc}
\toprule
\textbf{Metric} & \textbf{Mean $\pm$ SD} & \textbf{5th\,/\,95th \%} & \textbf{min\,/\,max} & \boldmath$\sigma / \mu$ \\
\midrule
R$L_2$ $\downarrow$ & $0.1470 \pm 0.0007$ & $0.1459$\,/\,$0.1483$ & $0.1454$\,/\,$0.1489$ & $0.50\%$ \\
SSIM   $\uparrow$   & $0.6720 \pm 0.0041$ & $0.6647$\,/\,$0.6779$ & $0.6590$\,/\,$0.6792$ & $0.61\%$ \\
PSDD   $\downarrow$ & $(3.70 \pm 0.46) \times 10^{-2}$ & $2.96 \times 10^{-2}$\,/\,$4.43 \times 10^{-2}$ & $2.79 \times 10^{-2}$\,/\,$4.97 \times 10^{-2}$ & $12.3\%$ \\
\bottomrule
\end{tabular}
\end{table}

The pointwise metrics (R$L_2$, SSIM) cluster within roughly half a percent of their mean across $100$ seeds: the digits cited for the distilled sCM student in Table~\ref{tab:smoke_metrics} are therefore stable to the random draw of $\boldsymbol{\epsilon}$ at the precision reported, and the comparison against the FM teacher is not sensitive to seed choice. PSDD has a substantially wider relative spread ($\sigma / \mu \approx 12\%$), as expected: it is a relative $L_2$ distance between one-dimensional radial spectra (Appendix~\ref{appendix:metrics}), and small fluctuations in the high-wavenumber tail produce a disproportionately large response in this metric. Even at the 95th-percentile worst case ($4.43 \times 10^{-2}$), the distilled student remains close to the multi-step diffusion baseline ($4.35 \times 10^{-2}$ at NFE\,$=30$ in Table~\ref{tab:smoke_metrics}) and only modestly above its own mean.

\clearpage

\section{Datasets}
\label{app:datasets}

We evaluate on three two-dimensional fluid-dynamics benchmarks spanning different resolutions, flow regimes, and dataset sizes. All datasets are split 90\%/10\% into training and validation; data statistics (per-channel mean and standard deviation) are pre-computed on the training split and used to standardise inputs to both teacher and student.

\subsection{2D Kolmogorov Flow}
\paragraph{Short description.}
12{,}800 snapshots of 2D periodic incompressible Kolmogorov Flow generated from the two-
dimensional incompressible Navier-Stokes equations with a sinusoidal forcing term at resolution $256\times256$. The coarse input is downsampled to $64\times64$ by stride-4 subsampling and nearest-neighbour upsampled back to $256\times256$ before inference, giving a $4\times$ super-resolution task.
\paragraph{Governing Equations.}
The 2D Kolmogorov Flow is simulated as a canonical benchmark for statistically stationary turbulence governed by the incompressible Navier-Stokes equations. The flow is described in the vorticity formulation of the 2D incompressible Navier-Stokes system,
\begin{equation}
    \frac{\partial \omega}{\partial t}
    + \mathbf{u}\cdot\nabla \omega
    = \nu \nabla^{2}\omega
    + \nabla \times \mathbf{F},
    \qquad
    \nabla \cdot \mathbf{u} = 0,
    \qquad
    \omega = \nabla \times \mathbf{u},
\end{equation}
where $\mathbf{u}(x,y,t)$ is the velocity field, $\omega(x,y,t)$ is the scalar vorticity, $\nu = 1/\mathrm{Re}$ is the kinematic viscosity, and $\mathbf{F}$ is a steady, large-scale sinusoidal body force
\begin{equation}
    \mathbf{F}(x,y) = 0.1\bigl[\sin(2\pi(x+y)) + \cos(2\pi(x+y))\bigr],
\end{equation}
that drives the flow into a statistically stationary turbulent regime. The system is solved on the periodic square $[0,1]\times[0,1]$ discretised on a $256\times256$ grid, with $\mathrm{Re}=1000$. Time integration uses a pseudo-spectral scheme with Crank--Nicolson stepping at $\Delta t = 10^{-3}\,\mathrm{s}$, with the nonlinear advective term computed in Fourier space and dealiased by the standard $2/3$-truncation rule. Each trajectory is initialised from a Gaussian random vorticity field with energy spectrum $E(k)\propto[1+(\tau k)^{2}]^{-\alpha}$ ($\alpha=2.5$, $\tau=7$), and snapshots are recorded at uniform temporal intervals over a horizon of $T=5000\,\mathrm{s}$ across multiple independent realisations to provide diverse, statistically stationary turbulent samples.

\subsection{Turbulent Channel Flow}
\paragraph{Short description.}
38{,}080 snapshots at resolution $192\times192$ at cross-section of 3D Turbulent Channel Flow. The low-resolution input is obtained by stride-3 sub-sampling to $64\times64$ and nearest-neighbour upsampled back to $192\times192$, preserving the anisotropic structure of near-wall turbulence.
\paragraph{Governing Equations.}
We use the dataset obtained from~\citet{xue2022synthetic}, in which the three-dimensional Turbulent Channel Flow is simulated by a large-eddy lattice Boltzmann method (LES-LBM). The LBM evolves a discrete-velocity distribution function $f_{i}(\mathbf{x},t)$ on a Cartesian lattice with velocity set $\{\mathbf{c}_{i}\}$ via the Bhatnagar--Gross--Krook (BGK) collision-streaming equation
\begin{equation}
    f_{i}(\mathbf{x}+\mathbf{c}_{i}\Delta t,\,t+\Delta t)
    - f_{i}(\mathbf{x},t)
    = -\frac{1}{\tau}\bigl[f_{i}(\mathbf{x},t) - f_{i}^{\mathrm{eq}}(\mathbf{x},t)\bigr],
\end{equation}
with the local equilibrium $f_{i}^{\mathrm{eq}}$ taken in its standard low-Mach polynomial form and the macroscopic density and momentum recovered as $\rho=\sum_{i}f_{i}$ and $\rho\mathbf{u}=\sum_{i}\mathbf{c}_{i}f_{i}$. The relaxation time $\tau$ is linked to an effective kinematic viscosity through $\nu_{\mathrm{eff}}=c_{s}^{2}(\tau-\tfrac{1}{2})\Delta t$, in which a Smagorinsky sub-grid eddy viscosity is embedded,
\begin{equation}
    \nu_{\mathrm{eff}} = \nu_{0} + \nu_{t},
    \qquad
    \nu_{t} = C_{\mathrm{smag}}\,\Delta^{2}\,\bar{S},
\end{equation}
where $\nu_{0}$ is the molecular viscosity, $\bar{S}$ is the magnitude of the filtered strain-rate tensor, and $\Delta$ is the lattice spacing. In the low-Mach, weakly-compressible limit, this scheme is consistent with the filtered incompressible Navier-Stokes equations,
\begin{equation}
    \frac{\partial \mathbf{u}}{\partial t}
    + \mathbf{u}\cdot\nabla \mathbf{u}
    = -\frac{1}{\rho}\nabla p
    + \nabla\cdot\bigl[(\nu_{0}+\nu_{t})(\nabla\mathbf{u}+\nabla\mathbf{u}^{\top})\bigr]
    + \mathbf{f},
    \qquad
    \nabla\cdot\mathbf{u} = 0,
\end{equation}
with $\mathbf{u}=(u_{x},u_{y},u_{z})$ along the streamwise, wall-normal and spanwise directions and $\mathbf{f}$ a uniform body force driving the channel. The simulation is carried out on a domain $L_{x}\times L_{y}\times L_{z}=768\times192\times192$ at friction Reynolds number $\mathrm{Re}_{\tau}=180$ (bulk $\mathrm{Re}=3250$), with periodic boundary conditions in the streamwise and spanwise directions and no-slip walls in the wall-normal direction. The flow is initialised from a quiescent state and perturbed by a cubic obstruction of size $20\times20\times100$ grid points placed at $x=192$ to trigger transition; after a long transient and removal of the obstruction the simulation is advanced for a further $50$ domain-through times to reach a statistically stationary turbulent state. The 2D fields used in this work are obtained by sampling the mid-plane cross-section at $x=384$ over $\sim100$ turnover times, yielding the $192\times192$ snapshots that serve as ground-truth high-fidelity data.

\subsection{2D Smoke Buoyancy}
\paragraph{Short description.} 173{,}824 snapshots at resolution $128\times128$, capturing buoyancy-driven scalar transport. The coarse input is downsampled to $32\times32$ by stride-4 subsampling and nearest-neighbour upsampled back to $128\times128$, giving a $4\times$ super-resolution factor. 
\paragraph{Governing Equations.} 
We use the dataset from \citet{zhou2024text2pde}, which is based on data from \citet{gupta2022towards} and was originally generated using the $\Phi_\mathrm{flow}$ (PhiFlow) solver~\cite{holl2024bf}. The simulations contain 56 timesteps, spanning $t=18$ to $t=102$, and are solved on a two-dimensional spatial domain, $x \in [0,32]$ and $y \in [0,32]$, at a resolution of $128 \times 128$. The data describes smoke driven by a buoyant force in a constrained box. The dynamics are governed by the incompressible Navier-Stokes equations, coupled with an advection equation for the smoke density $d$:
\begin{equation}
    \frac{\partial \mathbf{v}}{\partial t}
    + \mathbf{v} \cdot \nabla \mathbf{v}
    = \nu \nabla^2 \mathbf{v}
    - \frac{1}{\rho} \nabla p
    + \mathbf{f},
    \quad
    \nabla \cdot \mathbf{v} = 0.
\end{equation}
\begin{equation}
    \frac{\partial d}{\partial t}
    + \mathbf{v} \cdot \nabla d = 0.
\end{equation}

\clearpage

\section{Evaluation Metrics}\label{appendix:metrics}

We evaluate predicted 2D physical fields against ground truth using three complementary metrics: a pointwise relative error, a local structural similarity score, and a spectral discrepancy in Fourier space. Together they capture pixel-level fidelity, local geometric agreement, and the distribution of energy across spatial scales.

\subsection{Relative \texorpdfstring{$L_2$}{L2} Error}
\label{app:rel-l2}

Let $\hat{z}^{\,j}_{k}$ and $z^{\,j}_{k}$ denote the predicted and ground-truth states at step $k$ of the $j$-th test trajectory. The relative $L_2$ error is
\begin{equation}
\mathrm{rel.}\,L_{2}(k)
\;=\; \frac{1}{N}\sum_{j=1}^{N}
\frac{\bigl\lVert \hat{z}^{\,j}_{k}-z^{\,j}_{k}\bigr\rVert_{2}}
     {\bigl\lVert z^{\,j}_{k}\bigr\rVert_{2}},
\end{equation}
averaged over $N$ test trajectories. We report this quantity per step to characterize how error accumulates over the prediction horizon.

\subsection{Structural Similarity Index Measure (SSIM)}
\label{app:ssim}

To quantify local structural agreement between a reference field $x \in \mathbb{R}^{H \times W}$ and a prediction $y \in \mathbb{R}^{H \times W}$, we use SSIM \citep{wang2004image}, which compares local luminance, contrast, and structure within a sliding Gaussian window.

\paragraph{Local statistics.}
Let $\mathcal{H}(\cdot)$ denote convolution with a normalized 2D Gaussian kernel of window size $w \times w$ (with $w$ odd) and standard deviation $\sigma$, implemented as a separable 1D convolution with ``same'' padding. The local means, variances, and cross-covariance are
\begin{equation}
\mu_x = \mathcal{H}(x), \quad
\mu_y = \mathcal{H}(y), \quad
\sigma_x^2 = \mathcal{H}(x^2) - \mu_x^2, \quad
\sigma_y^2 = \mathcal{H}(y^2) - \mu_y^2, \quad
\sigma_{xy} = \mathcal{H}(xy) - \mu_x \mu_y.
\end{equation}

\paragraph{SSIM map and aggregation.}
The pointwise SSIM is
\begin{equation}
\mathrm{SSIM}(x,y)
\;=\;
\frac{(2\mu_x\mu_y + C_1)\,(2\sigma_{xy} + C_2)}
     {(\mu_x^2+\mu_y^2 + C_1)\,(\sigma_x^2+\sigma_y^2 + C_2)},
\end{equation}
with stability constants $C_1 = (K_1 L)^2$ and $C_2 = (K_2 L)^2$, using the standard choices $K_1 = 0.01$ and $K_2 = 0.03$. We report the spatial mean
\begin{equation}
\overline{\mathrm{SSIM}}(x,y) \;=\; \frac{1}{HW}\sum_{i=1}^{H}\sum_{j=1}^{W}\mathrm{SSIM}_{ij}(x,y),
\end{equation}
computed independently per sample for batched inputs of shape $(B, H, W, 1)$ and averaged over the batch.

\paragraph{Per-sample data range.}
Because our fields are unnormalized physical quantities whose magnitudes vary across samples, we set the data range adaptively as $L = \max(x) - \min(x)$ rather than using a fixed global value. In the degenerate case $L = 0$ (constant reference), we define $\overline{\mathrm{SSIM}} = 1$ when $x$ and $y$ agree to within numerical tolerance, and $0$ otherwise.

\subsection{Power Spectral Density Discrepancy (PSDD)}
\label{app:psd}

To assess whether predictions reproduce the correct distribution of energy across spatial scales---a property of central importance for turbulent and other multi-scale fields---we compare normalized 2D power spectra under an $\ell_1$ distance.

\paragraph{Mean-centering.}
We first remove the spatial mean of each sample to suppress the dominant DC component and emphasize fluctuations:
\begin{equation}
x \leftarrow x - \mathbb{E}_{i,j}[x], \qquad y \leftarrow y - \mathbb{E}_{i,j}[y],
\end{equation}
where $\mathbb{E}_{i,j}$ denotes averaging over spatial indices.

\paragraph{Centered Fourier transform.}
Let $\mathcal{F}\{\cdot\}$ denote the 2D DFT over spatial axes, applied per sample and channel. We centre the zero-frequency component via $\mathrm{fftshift}$:
\begin{equation}
\widetilde{F}_x \;=\; \mathrm{fftshift}\!\bigl(\mathcal{F}\{x\}\bigr), \qquad
\widetilde{F}_y \;=\; \mathrm{fftshift}\!\bigl(\mathcal{F}\{y\}\bigr).
\end{equation}

\paragraph{Overflow-safe normalized PSD.}
Naive squaring of $\widetilde{F}$ can overflow in single precision on high-dynamic-range scientific fields. We therefore rescale the real and imaginary parts by a per-sample factor
\begin{equation}
s \;=\; \max_{i,j,c}\,\max\!\bigl(|\Re(\widetilde{F}_{i,j,c})|,\,|\Im(\widetilde{F}_{i,j,c})|\bigr)
\end{equation}
prior to squaring, giving a numerically stable power spectrum
\begin{equation}
P \;=\; \left(\frac{\Re(\widetilde{F})}{\max(s,\varepsilon)}\right)^{\!2}
      + \left(\frac{\Im(\widetilde{F})}{\max(s,\varepsilon)}\right)^{\!2}.
\end{equation}
We then normalize $P$ to a probability distribution over frequency bins,
\begin{equation}
\widehat{P} \;=\; \frac{P}{\sum_{i,j,c} P_{i,j,c} + \varepsilon},
\end{equation}
which removes the residual amplitude dependence introduced by $s$ and isolates the \emph{relative} allocation of spectral energy.

\paragraph{Discrepancy.}
Given normalized PSDs $\widehat{P}^{(b)}_x$ and $\widehat{P}^{(b)}_y$ for the $b$-th sample, the PSDD is the per-bin $\ell_1$ distance averaged over space, channels, and the batch:
\begin{equation}
\mathcal{L}_{\mathrm{PSD}}(x,y)
\;=\; \frac{1}{B}\sum_{b=1}^{B}
\frac{1}{HWC}\sum_{i,j,c}
\left|\widehat{P}^{(b)}_{x,i,j,c} - \widehat{P}^{(b)}_{y,i,j,c}\right|.
\end{equation}
Lower values indicate closer agreement in spectral content between prediction and reference.

\clearpage

\section{Hyperparameters}
\label{app:hyperparameters}

This appendix lists the full set of hyperparameters used to produce the results in Section~\ref{sec:experiments}. All values correspond to the configurations of the FM teacher and sCM student checkpoints reported in the main paper, which we will release alongside the code.

All models are implemented in JAX/Flax NNX and trained on a single NVIDIA GPU per run, with Orbax checkpointing and an Adam optimiser ($\beta_1{=}0.9$, $\beta_2{=}0.999$, $\varepsilon{=}10^{-8}$). Inputs are normalised with per-dataset mean/standard deviation statistics computed from the training split; the train/validation split is $90\%/10\%$ across all datasets.

\subsection{Network architecture}
\label{app:hp_arch}

Both the FM teacher and the sCM student use the time-conditioned UNet of \cite{ho2020denoising} with sinusoidal time embeddings and GroupNorm. The low-resolution observation is concatenated channel-wise with the noised field at every forward pass; no auxiliary encoder is used. Single attention block at the $16{\times}16$ feature scale, dropout disabled ($p{=}0$), strided convolutions for resampling, single input/output channel for the scalar fields. Table~\ref{tab:hp_arch} lists the role-specific shape parameters; the spatial resolution follows the dataset.

\begin{table}[h]
\centering
\caption{UNet hyperparameters per role. $C$: base channel width (\texttt{latent\_dims}); $M$: per-level channel multipliers; $B$: residual blocks per level. The image resolution matches the target output of each dataset (Smoke Buoyancy: $128$; Turbulent Channel Flow: $192$; Kolmogorov Flow: $256$).}
\label{tab:hp_arch}
\small
\begin{tabular}{lccccc}
\toprule
Role & $C$ & $M$ & $B$ & Attention scale & Dropout \\
\midrule
FM teacher  & 96 & $(1,1,1,4)$ & 2 & $16{\times}16$ & 0.0 \\
sCM student & 96 & $(1,1,1,2)$ & 3 & $16{\times}16$ & 0.0 \\
\bottomrule
\end{tabular}
\end{table}

\subsection{Training}
\label{app:hp_training}

All runs use a fixed random seed of $42$, gradient-norm clipping at $1.0$, and the Adam optimiser. The two stages differ in their time sampler statistics, loss weighting and EMA usage. Table~\ref{tab:hp_optim} lists the optimiser-side hyperparameters for each (method, dataset) combination, and Table~\ref{tab:hp_method} lists the method-specific objective hyperparameters.

\begin{table}[h]
\centering
\caption{Optimiser-side training hyperparameters. ``Schedule'' refers to the learning-rate schedule; \emph{constant} keeps the listed value for the full run, \emph{cosine} decays linearly between the listed initial value and $10^{-5}$ over the indicated number of steps. Weight decay is applied only when listed.}
\label{tab:hp_optim}
\small
\setlength{\tabcolsep}{3pt}
\begin{tabular}{llccccc}
\toprule
Method & Dataset & Batch & Epochs & LR (init) & Schedule & Weight decay \\
\midrule
\multirow{3}{*}{FM teacher}
  & Smoke Buoyancy  & 64 & 150 & $1\mathrm{e}{-4}$ & cosine, $370{,}000$ steps & $1\mathrm{e}{-4}$ \\
  & Turbulent Channel Flow    & 16 & 150 & $1\mathrm{e}{-4}$ & constant                    & ---     \\
  & Kolmogorov Flow & 32 & 150 & $1\mathrm{e}{-4}$ & constant                    & $1\mathrm{e}{-4}$ \\
\midrule
\multirow{3}{*}{sCM student}
  & Smoke Buoyancy   & 16 & 150 & $2\mathrm{e}{-5}$ & constant & --- \\
  & Turbulent Channel Flow     & 16 & 150 & $2\mathrm{e}{-5}$ & constant & --- \\
  &  Kolmogorov Flow  & 16 & 150 & $2\mathrm{e}{-5}$ & constant & --- \\
\bottomrule
\end{tabular}
\end{table}

\begin{table}[h]
\centering
\caption{Method-specific objective hyperparameters. Time samples follow a logit-normal distribution $\log(t/(1-t))\sim\mathcal{N}(P_\mu, P_\sigma^2)$, clipped at $t_\varepsilon$ from the boundary. The FM teacher uses an adaptive loss weighting~\cite{karras2024analyzing} with norm $p{=}1$ and $\varepsilon_{\text{norm}}{=}0.01$; the sCM student uses an unweighted loss with TrigFlow tangent clipping~\cite{lu2024simplifying} and a linear warmup of the tangent term.}
\label{tab:hp_method}
\small
\begin{tabular}{lcc}
\toprule
Hyperparameter & FM teacher & sCM student \\
\midrule
$P_\mu$ (logit-normal mean)            & $-0.4$            & $-0.4$            \\
$P_\sigma$ (logit-normal std.)         & $1.0$             & $0.8$             \\
$t_\varepsilon$ (boundary clip)        & $10^{-3}$         & $10^{-6}$         \\
Adaptive loss weighting                & yes ($p{=}1$, $\varepsilon{=}0.01$) & no \\
EMA rate $\bar\alpha$                  & --- (unused at training)            & $0.98$ \\
Tangent clip                           & ---               & $1.0$             \\
Tangent-warmup iterations              & ---               & $500$             \\
\bottomrule
\end{tabular}
\end{table}

\subsection{Inference}
\label{app:hp_inference}

At inference time the conditioning input $\mathbf{x}_{\text{LR}}^{\uparrow}$ (the upsampled low-resolution field defined in Section~\ref{sec:problem}) is mixed with Gaussian noise to construct the initial state $\mathbf{z}_{t_{\text{start}}}$ via Eq.~\ref{eq:noise_injection}, after which the FM teacher integrates the learned velocity field with a fifth-order Runge--Kutta solver (RK5), while the sCM student performs a single forward pass. Table~\ref{tab:hp_inference} summarises the inference-time settings.

\begin{table}[h]
\centering
\caption{Inference settings. NFE: number of neural function evaluations per sample; these are dataset-independent.}
\label{tab:hp_inference}
\small
\begin{tabular}{lccc}
\toprule
Stage & Solver & NFE & $t_{\text{start}}$ \\
\midrule
FM teacher  & RK5    & 5 & $0.6$ \\
sCM student & ---    & 1 & $0.3$ \\
\bottomrule
\end{tabular}
\end{table}

\clearpage


\end{document}